\documentclass[lettersize,journal]{IEEEtran}
\usepackage{amsmath,amsfonts,amssymb,mathtools}
\usepackage{array}
\usepackage{graphicx}
\usepackage{cite}
\usepackage{booktabs}
\usepackage{float}
\usepackage{bm}
\usepackage{xcolor}
\usepackage{xr-hyper}
\usepackage{hyperref}
\usepackage{orcidlink}
\usepackage{capt-of}
\usepackage{multirow}
\usepackage{algorithm}
\usepackage{algpseudocode}
\hypersetup{hidelinks}

\graphicspath{{assets/}}

\newcommand{\method}{G-ray}
\newcommand{\authororcid}[1]{\hspace{-1.0mm}$^{\,\orcidlink{#1}}$}

\begin{document}
\title{G-ray: Ray-Level Relative Geometric\\Position Encoding in Multi-View Vision\\Transformers under Camera Heterogeneity}
\author{Shuo~Zhang\authororcid{0009-0004-5002-4577},~\IEEEmembership{Student~Member,~IEEE},
Xin~Su\authororcid{0000-0001-5901-8932},~\IEEEmembership{Member,~IEEE},
Wei~Wang\authororcid{0000-0003-0172-1582},~\IEEEmembership{Member,~IEEE},\\
Jun~Liu\authororcid{0000-0002-8943-079X},
Xinrui~Zeng\authororcid{0009-0002-4347-2545},
Yongsen~Chen\authororcid{0009-0007-7942-1826},
Chenjie~Wang\authororcid{0000-0001-9207-2076},
Guibo~Zhu\authororcid{0000-0001-8293-3952},\\
Jinqiao~Wang\authororcid{0000-0002-9118-2780},~\IEEEmembership{Member,~IEEE},
Bin~Luo\authororcid{0000-0002-3040-3500},~\IEEEmembership{Senior~Member,~IEEE},
and Liangpei~Zhang\authororcid{0000-0001-6890-3650},~\IEEEmembership{Fellow,~IEEE}%
\thanks{Shuo Zhang, Xin Su, Wei Wang, Jun Liu, Xinrui Zeng, Yongsen Chen, Bin Luo, and Liangpei Zhang are with Wuhan University, Wuhan, China.}%
\thanks{Guibo Zhu and Jinqiao Wang are with the Institute of Automation, Chinese Academy of Sciences, Beijing, China, and Wuhan AI Research, Wuhan, China.}%
\thanks{Chenjie Wang is with Rongyun Robot (Guizhou) Co., Ltd.}%
\thanks{Corresponding author: Bin Luo (e-mail: luob@whu.edu.cn).}}
\IEEEaftertitletext{%
  \vspace{-24pt}%
  \begin{minipage}{\textwidth}
    \centering
    \includegraphics[width=0.78\linewidth]{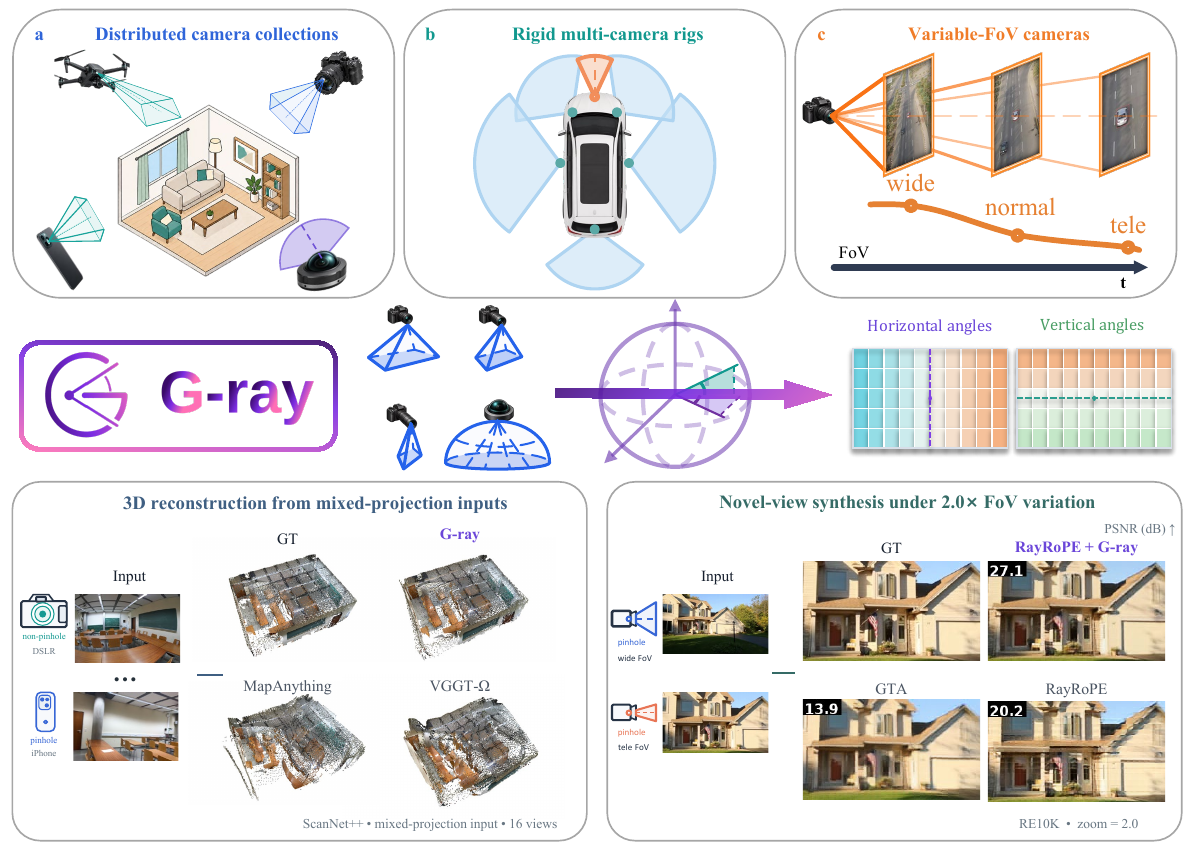}
    \captionof{figure}{Overview of \method{} under camera heterogeneity. Top, representative configurations include (a) distributed camera collections, (b) rigid multi-camera rigs, and (c) variable-FoV cameras. Middle, \method{} uses camera-local ray angles as positional coordinates across FoVs and projection models, illustrated by horizontal- and vertical-angle maps. Bottom, qualitative examples show 16-view 3D reconstruction on ScanNet++ from mixed non-pinhole DSLR and pinhole iPhone inputs, and NVS on RE10K under controlled FoV variation ($z{=}2.0$), with per-example PSNR values in dB. The top-row configurations illustrate application settings, not the capture setups of the bottom-row examples.}
    \label{fig:overview}
  \end{minipage}\par\vspace{6pt}%
}
\maketitle

\begin{abstract}
We study relative position encoding for multi-view vision Transformers under camera heterogeneity, including varying fields of view (FoVs) or projection models.
Existing rotary relative position encodings commonly use image-plane positional coordinates, producing projection-dependent relative phases and inconsistent geometric cues for cross-projection attention.
We introduce \textbf{\method{}}, a ray-level relative position encoding whose rotary phases are parameterized by camera-local ray angles.
The same camera-local ray pair induces the same relative phase across projections, providing projection-invariant positional consistency.
\method{} can be used directly or integrated with existing encodings, retaining complementary geometric cues without additional learned parameters.
We validate \method{} in three host encodings, RoPE, GTA, and RayRoPE, across 3D reconstruction and novel-view synthesis (NVS).
Across three heterogeneous 3D reconstruction benchmarks at 50 views, \method{} leads all six averaged metrics and reduces mean pointmap relative error by 45.8\% over MapAnything, with calibration supplied to both.
Trained exclusively on pinhole images, the 3D reconstruction model handles mixed pinhole and non-pinhole inputs without retraining and remains competitive on homogeneous pinhole 3D reconstruction protocols.
For NVS, GTA and RayRoPE improve with \method{} under joint viewpoint and FoV variation.
The project's webpage is available at \url{https://g-ray-project.github.io/}.
\end{abstract}


\section{Introduction}
\label{sec:intro}

\IEEEPARstart{M}{ulti-view} 3D perception is a foundational capability for embodied intelligence, augmented reality, autonomous driving, and mobile mapping, where a system needs to build spatially grounded representations of its surroundings from observations acquired across viewpoints~\cite{lin2024bip3d,pan2023ariadt,caesar2020nuscenes,yang2026uavff3d}.
To extend coverage, reduce blind spots, and capture complementary near- and far-field detail, practical systems often use a variety of camera configurations.
Three recurring configurations illustrate this diversity (Fig.~\ref{fig:overview}(a)--(c)).
\emph{Distributed camera collections} combine independently acquired views, potentially mixing devices, FoVs, and projection models~\cite{wang2013surveillance,natarajan2015multicamera,li2018megadepth,schops2017eth3d,yeshwanth2023scannetpp,zhou2024coped,hou2025agcdrive}.
\emph{Rigid multi-camera rigs} maintain fixed relative camera poses while combining different FoVs or projection models~\cite{pan2023ariadt,caesar2020nuscenes,xiao2021pandaset,zhou2026xlens}.
\emph{Variable-FoV cameras}, including continuous-zoom and pan-tilt-zoom units, vary their angular coverage over time~\cite{wang2013surveillance,natarajan2015multicamera,amosa2023multicamera}.
We use \emph{camera heterogeneity} to describe such configurations, where imaging geometry such as FoV or projection model varies across views, in contrast to \emph{homogeneous} input whose views share a single imaging geometry~\cite{ganesan2026unidac,guo2026panovggt}.
Any perception model built on heterogeneous input must therefore integrate views that differ not only in where they observe the scene but also in how 3D viewing rays are mapped onto image locations.

Classical geometric formulations use calibrated camera models~\cite{zhang2000calib,rameau2022mccalib} to convert image measurements from different projections into viewing rays.
This ray-based representation thus can accommodate camera heterogeneity during cross-view geometric reasoning~\cite{schonberger2016sfm,yao2018mvsnet}.
The limitations of these formulations lie elsewhere.
They rely on computationally expensive iterative optimization, and their accuracy can degrade under weak texture, dynamic content, and sparse views~\cite{wang2025vggt,wang2026vggtomega}.
This cost and brittleness motivated the shift to feed-forward multi-view vision Transformers, which replace staged geometric optimization with a single learned system~\cite{wang2024dust3r,leroy2024mast3r,cabon2025mustr3r,wang2025vggt,yang2025fast3r,wang2025pi3,lin2026da3,keetha2025mapanything,jin2024lvsm}.
The shift, however, also discarded the explicit camera model.
Early designs in this family inherit the positional encoding of language and single-image vision Transformers, which tells a token where it sits within its own image but nothing about the camera that produced it~\cite{vaswani2017attention,dosovitskiy2021vit,su2021roformer,heo2024rope2d,liu2021swin,wu2021irpe,chu2021cpe}.
The problem of camera heterogeneity is therefore not resolved but dropped from the formulation.

\begin{figure}[!t]
  \centering
  \includegraphics[width=\columnwidth]{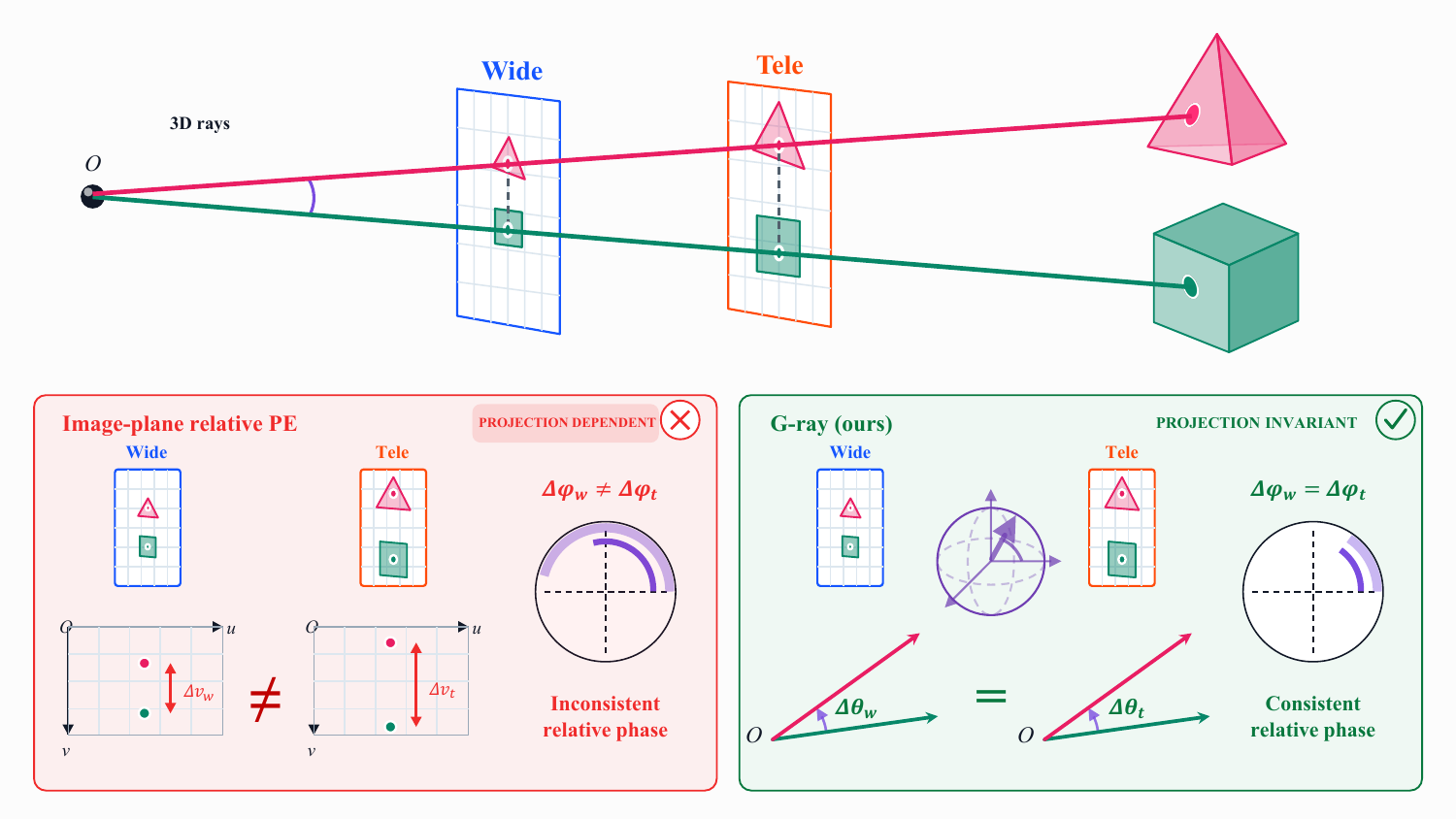}
  \caption{Cross-projection phase mismatch and ray-angle consistency. Wide- and tele-FoV projections map the same camera-local ray pair to different image-plane displacements, yielding unequal relative phases (left). \method{} instead uses ray-angle coordinate differences, so the same ray pair induces the same relative phase across projections (right).}
  \label{fig:image_plane_problem}
\end{figure}

Subsequent feed-forward methods restore camera awareness through two families of positional encodings.
Absolute camera encodings, including Pl\"ucker-ray embeddings~\cite{sitzmann2021light,jin2024lvsm,yin2026raype}, 3D point positional encoding~\cite{shu2023p3dpe}, and CamRay tokens~\cite{keetha2025mapanything}, attach viewing geometry to individual tokens by addition or concatenation.
Because geometry is encoded per token rather than per token pair, cross-view relative structure must be inferred indirectly from absolute features, without an explicit guarantee of consistent positional behavior across viewpoints~\cite{su2021roformer,miyato2023gta,li2025prope}.
Relative camera encodings instead incorporate camera parameters into token-pair interactions~\cite{kenney2026dppe,zhang2026ucpe}.
GTA and PRoPE combine camera-transform components with rotary image-grid subspaces~\cite{miyato2023gta,li2025prope}, while projection-based rotary encodings such as RayRoPE, URoPE, and CRePE obtain rotary coordinates by projecting geometric quantities into a query camera~\cite{wu2026rayrope,xie2026urope,jin2026crepe}.
Although these approaches introduce geometric information into relative attention, their grid-index or projected-pixel components still form relative phases from image-plane positional coordinates.
In the lower-left panel of Fig.~\ref{fig:image_plane_problem}, wide- and tele-FoV projections map the same pair of viewing rays to different image-plane displacements.
These displacements induce unequal relative phases even though the underlying ray relation is unchanged, making the positional treatment projection-dependent.
We refer to this effect as \emph{cross-projection phase mismatch}, a source of inconsistent geometric cues for cross-projection attention.

We address this mismatch with \textbf{\method{}}, a ray-level relative position encoding that lifts the positional coordinate from the image plane to \emph{ray-angle space}.
\method{} converts each positional sample to a camera-local ray using a view-specific image-to-ray map and represents it by horizontal and vertical angles $(\theta_x, \theta_y)$.
This conversion expresses viewing direction in radians, giving the positional coordinates a shared physical meaning across projections.
In the lower-right panel of Fig.~\ref{fig:image_plane_problem}, the wide- and tele-FoV observations recover the same angular separation for the same ray pair and therefore the same relative phase.
\method{} thus provides projection-invariant positional consistency, with relative phases determined by camera-local ray-angle coordinate differences rather than projection-dependent image-plane displacements.

\method{} defines ray-angle rotary transformations that can serve directly as a relative position encoding or be integrated into existing geometric encodings (Fig.~\ref{fig:replacement}).
Integration replaces the host's grid-index or projected pixel coordinates with ray-angle pairs while retaining its complementary geometric components, such as depth, disparity, pose, and camera-center information.
The image-to-ray map $\mathcal{U}_c$ is derived from intrinsics for pinhole views and supplied by a calibrated projection model or unit-ray map for non-pinhole views.
An external estimator can also supply this geometry when calibration is unavailable.
The conversion itself introduces no learned parameters and requires neither token depth nor camera extrinsics, although the host encoding may still use both.

We test \method{} in 3D reconstruction and novel-view synthesis (NVS) using three host encodings.
For 3D reconstruction, we use \method{} directly as the ray-angle rotary position encoding in a ViT-based model, replacing grid-index RoPE.
For NVS, we replace either GTA's image-grid coordinates~\cite{miyato2023gta} or RayRoPE's query-camera projected pixels~\cite{wu2026rayrope} within the same LVSM backbone~\cite{jin2024lvsm}, retaining each host encoding's remaining geometric components.
These instantiations test whether shared ray-angle coordinates benefit both grid-based and projection-based rotary formulations across tasks under camera heterogeneity.
Notably, the 3D reconstruction model trained exclusively on pinhole inputs handles mixed pinhole and non-pinhole inputs without retraining.

On heterogeneous 3D reconstruction across PandaSet, ETH3D under controlled FoV variation, and mixed-projection ScanNet++, \method{} leads all six dataset-averaged metrics at 50 views among methods evaluated on all three benchmarks.
On mixed-projection ScanNet++ at 50 views, \method{} reduces both pointmap and ray-depth relative errors by about 74\% compared with MapAnything when both use supplied calibration, while increasing point inliers from 26.2\% to 75.1\%.
In NVS, integrating \method{} into GTA improves all three main Objaverse regimes and substantially improves RE10K results under controlled FoV variation.
For RayRoPE, the largest gain among the three main Objaverse regimes occurs at novel viewpoints with additional FoV change.
Further experiments show more gradual degradation above the training resolution, effective 3D reconstruction with AnyCalib-estimated calibration~\cite{tiradogarin2025anycalib}, and competitive accuracy on homogeneous pinhole 3D reconstruction protocols.

Our contributions are as follows.
\begin{itemize}
\item We introduce \method{}, a direct ray-level angular encoding that gives relative phase a shared geometric interpretation and provides projection-invariant positional consistency. This encoding addresses the cross-projection phase mismatch we identify in rotary relative position encodings based on image-plane coordinates (\S\ref{sec:method}).
\item We formulate \method{} for direct use as a ray-angle rotary position encoding and for integration with existing geometric encodings. Instantiations in RoPE, GTA, and RayRoPE cover grid-based and projection-based formulations while retaining each host encoding's complementary geometry and adding no learned parameters (\S\ref{subsec:replacement}).
\item We validate \method{} in 3D reconstruction and NVS under camera heterogeneity. 3D reconstruction evaluation spans distributed camera collections (ScanNet++), rigid multi-camera rigs (PandaSet), and variable-FoV inputs (ETH3D under controlled FoV variation). \method{} leads all six dataset-averaged metrics at 50 views and achieves the lowest ray-direction error in each setting. In NVS, both GTA and RayRoPE improve with \method{} under joint viewpoint and FoV variation (\S\ref{sec:experiments}).
\item We demonstrate pinhole-to-non-pinhole transfer in 3D reconstruction, applying the same model trained on pinhole inputs to heterogeneous views at inference without retraining. We further evaluate \method{} for resolution extrapolation and 3D reconstruction with calibration estimated from the input images (\S\ref{sec:experiments}).
\end{itemize}

\section{Related Work}
\label{sec:related}

\subsection{Camera Heterogeneity in 3D Perception}

Camera heterogeneity is not confined to a particular task.
It arises when input views use different FoVs or projection models, whether acquired by multiple devices or a single variable-FoV camera.
\emph{Distributed camera collections} combine independently acquired views that may differ in device, FoV, and projection model, as in crowd-sourced 3D reconstruction and scene understanding~\cite{li2018megadepth,schops2017eth3d,yeshwanth2023scannetpp}.
\emph{Rigid multi-camera rigs} maintain fixed relative camera poses while combining different FoVs or projection models, as in autonomous driving and egocentric sensing~\cite{caesar2020nuscenes,xiao2021pandaset,pan2023ariadt}.
\emph{Variable-FoV cameras}, including continuous-zoom and pan-tilt-zoom units, change their angular coverage over time~\cite{wang2013surveillance,natarajan2015multicamera,amosa2023multicamera}.
Surveillance tasks involve detection and tracking across camera networks with overlapping or non-overlapping FoVs, using static, pan-tilt-zoom, and omnidirectional cameras~\cite{song2013sparsecamera,natarajan2015multicamera,amosa2023multicamera}.
We use \emph{camera heterogeneity} for variation in imaging geometry within one multi-view input, which may also involve viewpoint changes.

Recent learning-based systems address selected portions of this design space.
PanoVGGT reconstructs panoramic imagery~\cite{guo2026panovggt}, UniDAC estimates metric depth across camera models~\cite{ganesan2026unidac}, X-Lens combines ray maps with projection-aware features for mixed pinhole and fisheye rigs~\cite{zhou2026xlens}, and UAVFF3D studies feed-forward 3D reconstruction under changes in aerial-view geometry~\cite{yang2026uavff3d}.
These works demonstrate the value of projection-aware modeling, but each is tied to a particular task, camera family, or complete system.
\method{} instead asks whether heterogeneous views can share one relative positional coordinate, and evaluates that coordinate in both 3D reconstruction and NVS.

\subsection{Classical and Feed-Forward Visual Geometry}

Classical image-based geometry resolves the view-specific projection explicitly when combining observations.
Structure-from-motion and multi-view stereo estimate correspondences, cameras, and scene structure through triangulation and geometric optimization~\cite{schonberger2016sfm,yao2018mvsnet}.
Per-scene NVS methods similarly cast calibrated pixels into camera rays and optimize a radiance field or Gaussian scene representation~\cite{mildenhall2020nerf,kerbl2023gaussians}.
Calibrated multi-camera detection and tracking systems also use projection geometry to associate observations across views~\cite{natarajan2015multicamera,amosa2023multicamera}.
When an appropriate camera model is supplied, these formulations can unproject pinhole, fisheye, or zoom observations into a common geometric space.
Their generality, however, often depends on reliable calibration and correspondence, staged processing, iterative optimization, or per-scene fitting.

Feed-forward models replace much of this processing with learned cross-view aggregation.
For 3D reconstruction, DUSt3R introduced pairwise pointmap regression~\cite{wang2024dust3r}, MASt3R added dense matching features~\cite{leroy2024mast3r}, and MUSt3R generalized the formulation to multiple views~\cite{cabon2025mustr3r}.
VGGT jointly predicts cameras, depth, pointmaps, and tracks~\cite{wang2025vggt}, while VGGT-$\Omega$, $\pi^3$, Fast3R, Depth Anything~3, and MapAnything extend robustness, scale, equivariance, or optional geometric conditioning~\cite{wang2026vggtomega,wang2025pi3,yang2025fast3r,lin2026da3,keetha2025mapanything}.
For NVS, pixelSplat and MVSplat predict Gaussian scene representations~\cite{charatan2024pixelsplat,chen2024mvsplat}, whereas LVSM directly maps context views to target-view appearance with minimal explicit 3D inductive bias~\cite{jin2024lvsm}.
Multi-view Transformers have also advanced camera-only 3D detection and temporal tracking through 3D queries, position-aware image features, bird's-eye-view aggregation, and object-centric temporal propagation~\cite{wang2022detr3d,liu2022petr,li2022bevformer,wang2023streampetr}.
These models make visual geometry substantially faster and more unified, but most are trained and evaluated with a shared perspective projection or a homogeneous camera rig.
This leaves open which coordinate should parameterize relative position when one attention layer receives tokens formed under different projection geometries.

\subsection{Positional Encoding for Vision Transformers}

Self-attention is permutation equivariant and therefore requires an explicit source of positional structure~\cite{vaswani2017attention,dosovitskiy2021vit}.
Absolute encodings associate each token with a learned or fixed positional feature.
Relative encodings instead parameterize token-pair relations, including window-relative biases and the directional, context-dependent offsets of iRPE~\cite{liu2021swin,wu2021irpe}.
Conditional positional encoding generates local position cues from the input neighborhood and improves extrapolation to unseen image sizes~\cite{chu2021cpe}, and ViTAR trains with perturbed positions so that a model tested above its training resolution encounters positional values it has effectively already seen~\cite{fan2024vitar}.
RoPE applies position-dependent orthogonal rotations so that query-key products depend on relative displacement~\cite{su2021roformer}, and its two-dimensional variants extend this mechanism to image lattices~\cite{heo2024rope2d}.

Most image positional encodings are designed around a regular two-dimensional lattice.
They improve locality, translation structure, or resolution extrapolation while presuming that positions from different images share the same image-plane metric.
Multi-view rotary coordinates inherit this assumption through either token-grid indices, as in standard 2D RoPE and the spatial subspace of GTA, or pixels obtained by projection into a query camera, as in RayRoPE.
The two sources differ computationally, but both remain tied to projection-specific image coordinates.
\method{} changes the coordinate supplied to the relative rotation from an image-plane pair to camera-local horizontal and vertical ray angles.

\subsection{Camera-Aware Encoding in Multi-View Transformers}

Camera-aware encoding in multi-view Transformers falls into two families that parallel the absolute-versus-relative distinction in standard positional encoding.

\paragraph{Absolute camera encodings}
attach viewing geometry to individual tokens through addition, concatenation, or adapter-based conditioning.
Representative designs encode 3D points or Pl\"ucker rays as token-aligned features~\cite{shu2023p3dpe,sitzmann2021light}, MapAnything represents optional calibration through CamRay inputs~\cite{keetha2025mapanything}, the released Pi3X extension of $\pi^3$ accepts optional poses, intrinsics or rays, and depth~\cite{wang2025pi3},\footnote{\href{https://github.com/yyfz/Pi3}{Official Pi3X release documentation}.}
WorldMirror incorporates camera and depth priors through any-prior prompting~\cite{liu2026worldmirror}, and OmniVGGT uses a GeoAdapter to inject arbitrary subsets of geometric modalities~\cite{peng2025omnivggt}.
Camera-aware generation methods similarly encode per-token Pl\"ucker rays, depth-aware projected paths, or relative ray and orientation cues~\cite{yin2026raype,jin2026crepe,zhang2026ucpe}.
Because geometry is carried per token rather than expressed in pairwise attention, cross-view relative structure must be inferred indirectly from individual token features, without an explicit guarantee of consistent positional behavior across viewpoints.

\paragraph{Relative camera encodings}
incorporate camera parameters into the token-pair interaction itself.
GTA combines a camera-transformation subspace with a regular image-grid subspace~\cite{miyato2023gta}, PRoPE models relative viewing-frustum relationships~\cite{li2025prope}, and DPPE factorizes camera transformations for scalable multi-view attention~\cite{kenney2026dppe}.
RayRoPE projects a depth-parameterized point on each ray into a query camera and includes its query-frame pixel coordinate and disparity in the rotary position~\cite{wu2026rayrope}.
URoPE lifts key-view pixels to multiple depth anchors, projects them into the query view, and assigns the resulting positions to attention heads~\cite{xie2026urope}.
Although these approaches introduce geometric information into relative attention, their image-plane rotary components remain projection-dependent, so cross-projection phase mismatch persists.

\method{} isolates a narrower common component in this landscape.
It replaces only the projection-dependent coordinate pair used by a relative rotary encoding with camera-local ray angles, while retaining complementary ray features, pose transformations, disparity, uncertainty, or depth-dependent branches of the surrounding method.

\section{Ray-Level Relative Geometric Position Encoding}
\label{sec:method}

\begin{figure*}[!t]
  \centering
  \includegraphics[width=\textwidth]{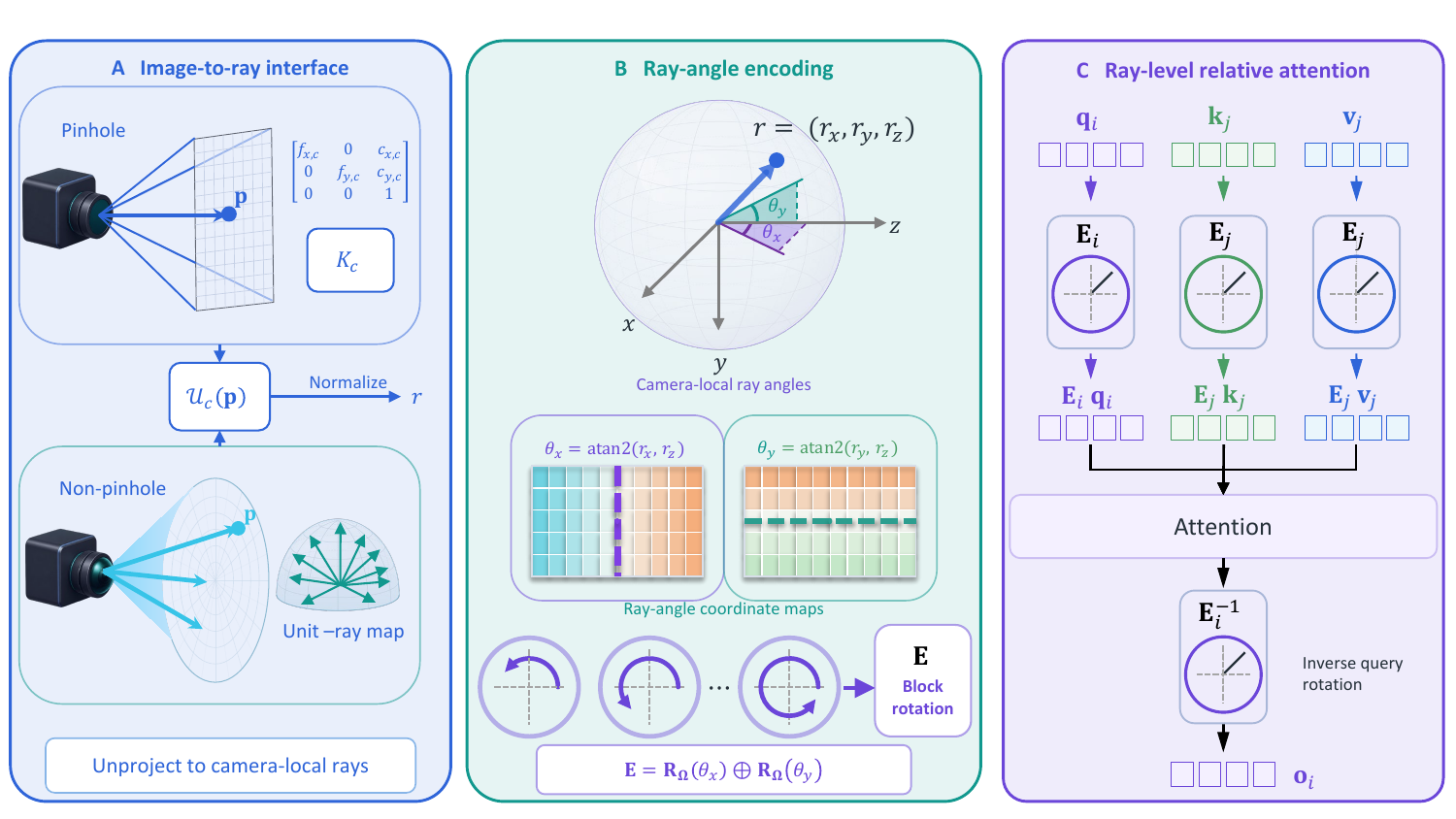}
  \caption{\method{} from image-to-ray geometry to ray-level relative attention. (A) View-specific unprojection and normalization convert an image-plane positional sample to a camera-local unit ray, using pinhole intrinsics or a non-pinhole unit-ray map. (B) Horizontal and vertical angles, computed with $\operatorname{atan2}$, form ray-angle coordinate maps and parameterize multi-frequency block rotations. (C) Queries, keys, and values are rotated before attention, followed by the inverse query rotation on the output.}
  \label{fig:method}
\end{figure*}

\method{} replaces image-plane positional coordinates with camera-local ray-angle coordinates while retaining the host encoding's complementary geometry and surrounding architecture.
Fig.~\ref{fig:method} organizes the resulting computation into three stages.
An image-to-ray interface maps each image-plane positional sample to a camera-local viewing ray, in closed form for a pinhole camera and through a per-pixel unit-ray map for a non-pinhole camera (Fig.~\ref{fig:method}A).
Horizontal and vertical ray angles form patch-aligned coordinate maps and parameterize each token's multi-frequency block rotation (Fig.~\ref{fig:method}B).
The rotations are applied to queries, keys, and values before attention, followed by the inverse query rotation on the output (Fig.~\ref{fig:method}C).
The image-to-ray conversion is view-specific, while the ray-angle parameterization and rotary construction are shared across projections.
Sec.~\ref{subsec:problem} identifies the positional variable we replace, Sec.~\ref{subsec:angle} defines the ray-angle coordinate, and Sec.~\ref{subsec:operator} defines the attention computation.
Sec.~\ref{subsec:properties} explains the resulting properties and calibration requirement, with a qualitative diagnostic of positional consistency.
Sec.~\ref{subsec:replacement} describes the replacement in three host encodings.

\subsection{Problem Formulation}
\label{subsec:problem}

We consider a rotary encoding parameterized by a two-dimensional positional coordinate.
Let a multi-view input contain images $\{\mathbf{I}_c\}_{c=1}^{C}$, and let token $i$ of view $c$ carry a feature $\mathbf{x}_{c,i}$ and a positional coordinate $\boldsymbol{\xi}_{c,i}\in\mathbb{R}^{2}$.
The encoding assigns each token an orthogonal transformation $\mathbf{E}(\boldsymbol{\xi}_{c,i})$, yielding the query-key interaction
\begin{equation}
\left(\mathbf{E}(\boldsymbol{\xi}_{c,i})\mathbf{q}_{c,i}\right)^{\!\top}
\left(\mathbf{E}(\boldsymbol{\xi}_{d,j})\mathbf{k}_{d,j}\right)
=
\mathbf{q}_{c,i}^{\top}
\mathbf{E}(\boldsymbol{\xi}_{c,i})^{\top}
\mathbf{E}(\boldsymbol{\xi}_{d,j})
\mathbf{k}_{d,j}.
\label{eq:rotary_phase}
\end{equation}
The positional effect of this rotary component depends on the coordinate pair $(\boldsymbol{\xi}_{c,i},\boldsymbol{\xi}_{d,j})$.
We study which coordinates should parameterize it under camera heterogeneity.

Existing encodings supply $\boldsymbol{\xi}$ as token-grid indices or query-camera projected pixel coordinates, both defined on the image plane.
For a zero-skew pinhole camera, a ray with horizontal angle $\theta_x$ projects to $u=c_x+f_x\tan\theta_x$.
Consequently, equal pixel displacements can represent different ray-angle differences, both across focal lengths and between the center and periphery of one image.
Non-pinhole cameras, such as fisheye cameras, map ray directions to image coordinates differently, so this correspondence also depends on the projection model.
Relative phases derived from these projection-dependent displacements therefore provide inconsistent geometric cues for cross-projection attention.

\method{} instead expresses position through viewing direction, using angular units shared across projections.
We therefore replace $\boldsymbol{\xi}_{c,i}$ with a camera-local ray-angle coordinate $\boldsymbol{\theta}_{c,i}$ before the rotary phase is formed.
The conversion uses the view-specific image-to-ray map
\begin{equation}
\mathcal{U}_c: \Omega_c \rightarrow \mathbb{R}^{3}\setminus\{\mathbf{0}\},
\label{eq:image_to_ray_map}
\end{equation}
which sends a location in the valid image domain $\Omega_c$ to a nonzero camera-frame direction.
This reparameterization removes projection dependence from the positional phase.
The host encoding's complementary geometry, including pose conditioning, depth prediction, and camera-transform subspaces, is retained (Sec.~\ref{subsec:replacement}).

\subsection{Ray-Angle Coordinates}
\label{subsec:angle}

\method{} turns every image-plane positional sample into a pair of camera-local ray angles.
Let $\mathbf{p}_{c,i}=(u_{c,i},v_{c,i},1)^{\top}$ be a positional sample of token $i$ in view $c$.
Unprojection followed by normalization gives the unit viewing ray
\begin{equation}
\mathbf{r}_{c,i}
=
\frac{\mathcal{U}_c(\mathbf{p}_{c,i})}
{\|\mathcal{U}_c(\mathbf{p}_{c,i})\|_2}
=
(r_{x,c,i},\,r_{y,c,i},\,r_{z,c,i})^{\top},
\label{eq:camera_ray}
\end{equation}
which we parameterize by its horizontal and vertical angles
\begin{align}
\theta_{x,c,i} &= \operatorname{atan2}\left(r_{x,c,i},\,r_{z,c,i}\right), \label{eq:thetax} \\
\theta_{y,c,i} &= \operatorname{atan2}\left(r_{y,c,i},\,r_{z,c,i}\right). \label{eq:thetay}
\end{align}
The ray-angle coordinate of the sample is $\boldsymbol{\theta}_{c,i}=(\theta_{x,c,i},\theta_{y,c,i})$, shown as the two angle maps of Fig.~\ref{fig:method}B.
The two-argument arctangent preserves quadrant information, with principal values in $(-\pi,\pi]$.
The occupied angular range depends on the view's FoV and projection model.
For forward-facing rays, both angles lie in $(-\pi/2,\pi/2)$.

The projection model enters through $\mathcal{U}_c$, which is instantiated differently for pinhole and non-pinhole projections (Fig.~\ref{fig:method}A).
A pinhole camera with intrinsics $\mathbf{K}_c$ admits the closed form $\mathcal{U}_c(\mathbf{p})=\mathbf{K}_c^{-1}\mathbf{p}$.
For zero skew, \eqref{eq:thetax} and \eqref{eq:thetay} reduce to $\theta_x=\operatorname{atan2}(u-c_x,\,f_x)$ and $\theta_y=\operatorname{atan2}(v-c_y,\,f_y)$.
For non-pinhole cameras, we represent $\mathcal{U}_c$ as a per-pixel unit-ray map.
This map can be computed from a parametric projection model, supplied by factory calibration, or predicted by an external estimator.
Our mixed-projection evaluation uses the four-coefficient fisheye model of the ScanNet++ DSLR stream.
This model maps a ray at polar angle $\psi$ from the optical axis to the normalized radius
\begin{equation}
\rho(\psi)
=
\psi\left(1 + k_1\psi^{2} + k_2\psi^{4} + k_3\psi^{6} + k_4\psi^{8}\right),
\label{eq:fisheye}
\end{equation}
which we numerically invert to obtain $\psi$ and reconstruct the ray (Appendix~\ref{app:image_to_ray_maps}).
The same angular parameterization in \eqref{eq:thetax} and \eqref{eq:thetay} then applies to both pinhole and non-pinhole views.

\subsection{Relative Angular Attention}
\label{subsec:operator}

The ray-angle coordinate enters attention through the standard rotary construction.
For head dimension $d_h$, \method{} builds the block-diagonal transformation
\begin{equation}
\mathbf{E}_{c,i}
=
\mathbf{R}_{\boldsymbol{\Omega}}(\theta_{x,c,i})
\oplus
\mathbf{R}_{\boldsymbol{\Omega}}(\theta_{y,c,i}),
\label{eq:rotation}
\end{equation}
where $\oplus$ denotes block-diagonal concatenation and $\mathbf{R}_{\boldsymbol{\Omega}}(\theta)$ applies a bank of planar rotations to disjoint channel pairs.
We parameterize a geometric bank of angular frequencies as
\begin{equation}
\omega_k
=
\alpha\,\beta^{-k/n_{\Omega}},
\label{eq:frequency}
\end{equation}
in which $\alpha$ is a frequency scale, $\beta$ is a frequency base, and $n_{\Omega}$ is the number of frequency components in one angular subspace, with $k=0,\ldots,n_{\Omega}-1$.
For axis $a\in\{x,y\}$, the rotary phase of token $i$ in view $c$ at frequency $k$ is
\begin{equation}
\varphi^{a}_{c,i,k}=\omega_k\theta_{a,c,i}.
\label{eq:angular_phase}
\end{equation}
Each channel pair is rotated using $\cos\varphi^{a}_{c,i,k}$ and $\sin\varphi^{a}_{c,i,k}$, so the positional coordinate and frequency jointly determine the phase supplied to attention.
Ray angles and image-plane coordinates have different numerical scales, so $\alpha$ controls the range of rotation phases.
Sec.~\ref{subsec:ablation_nvs} ablates the frequency scale.

\method{} transforms queries, keys, and values and applies the inverse query transformation to the attention output
\begin{align}
\widetilde{\mathbf{q}}_{c,i} &= \mathbf{E}_{c,i}\mathbf{q}_{c,i}, \qquad
\widetilde{\mathbf{k}}_{d,j} = \mathbf{E}_{d,j}\mathbf{k}_{d,j}, \\
\widetilde{\mathbf{v}}_{d,j} &= \mathbf{E}_{d,j}\mathbf{v}_{d,j}, \label{eq:qkv} \\
\widetilde{\mathbf{q}}_{c,i}^{\top}\widetilde{\mathbf{k}}_{d,j}
&=
\mathbf{q}_{c,i}^{\top}\mathbf{E}_{c,i}^{\top}\mathbf{E}_{d,j}\mathbf{k}_{d,j}, \label{eq:relative} \\
\mathbf{o}_{c,i}
&=
\mathbf{E}_{c,i}^{-1}
\operatorname{Attention}\!\left(
\mathbf{E}_{c,i}\mathbf{q}_{c,i},
\{\mathbf{E}_{d,j}\mathbf{k}_{d,j},\mathbf{E}_{d,j}\mathbf{v}_{d,j}\}_{d,j}
\right). \label{eq:output}
\end{align}
Because planar rotations compose by angle addition, the transpose-query rotation in \eqref{eq:relative} subtracts the query phase from the key phase.
The relative phase for each axis and frequency is therefore
\begin{equation}
\begin{aligned}
\Delta\varphi^{a}_{c i,d j,k}
&=\varphi^{a}_{d,j,k}-\varphi^{a}_{c,i,k} \\
&=\omega_k\bigl(\theta_{a,d,j}-\theta_{a,c,i}\bigr).
\end{aligned}
\label{eq:relative_angular_phase}
\end{equation}
For the same ray pair, Fig.~\ref{fig:image_plane_problem} uses $\Delta\varphi_w$ and $\Delta\varphi_t$ to denote this relative phase under wide- and tele-FoV projections.
Thus, $\mathbf{E}_{c,i}^{\top}\mathbf{E}_{d,j}$ is parameterized by horizontal and vertical angular offsets rather than projection-dependent pixel offsets.
These offsets compare camera-local coordinates, while any complementary camera-pose transformations remain part of the host encoding.
The formulation in \eqref{eq:qkv} to~\eqref{eq:output} supports within-view and cross-view attention in encoder and encoder-decoder architectures.

\begin{figure}[!t]
  \centering
  \includegraphics[width=\columnwidth]{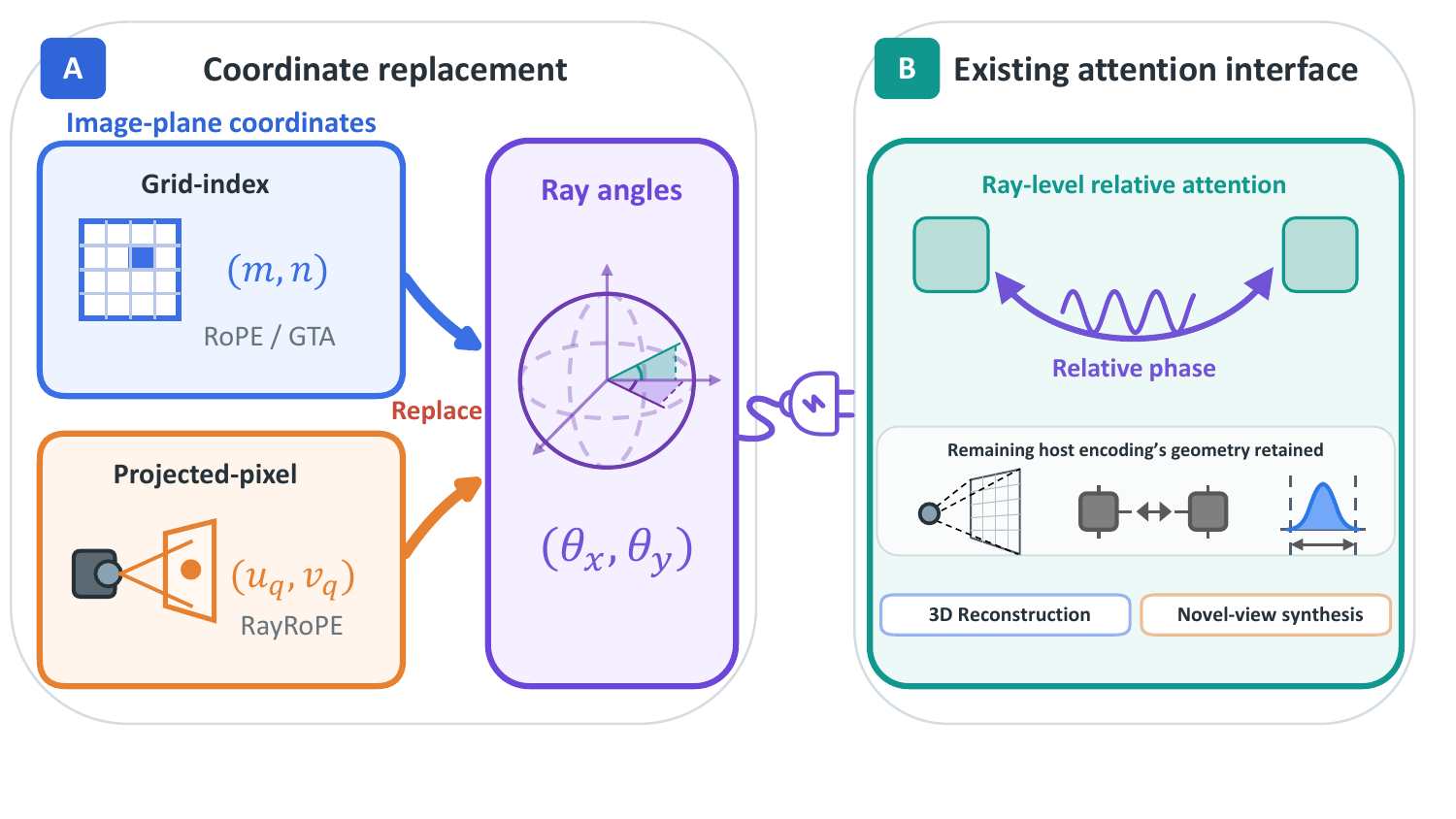}
  \caption{Integrating \method{} with existing rotary encodings. (A) Grid-index coordinates in RoPE or GTA, or query-camera projected-pixel coordinates in RayRoPE, are replaced by camera-local ray-angle coordinates. (B) Ray-angle differences determine relative phases within the existing attention interface, while the host encoding's complementary geometric components are retained. Each variant is trained independently.}
  \label{fig:replacement}
\end{figure}

\subsection{Properties and Requirements}
\label{subsec:properties}

Replacing the positional coordinate changes what the rotary phase measures.
We describe the resulting positional consistency and coordinate range, together with their calibration requirement.

\subsubsection{Projection-Invariant Positional Consistency}
\label{subsec:attn_diag}

An angular increment has a shared physical meaning across projections.
The same camera-local ray-angle coordinate difference therefore produces the same relative phase, whereas equal image-plane displacements need not represent equal changes in viewing direction.

\begin{figure*}[!t]
  \centering
  \includegraphics[width=\textwidth]{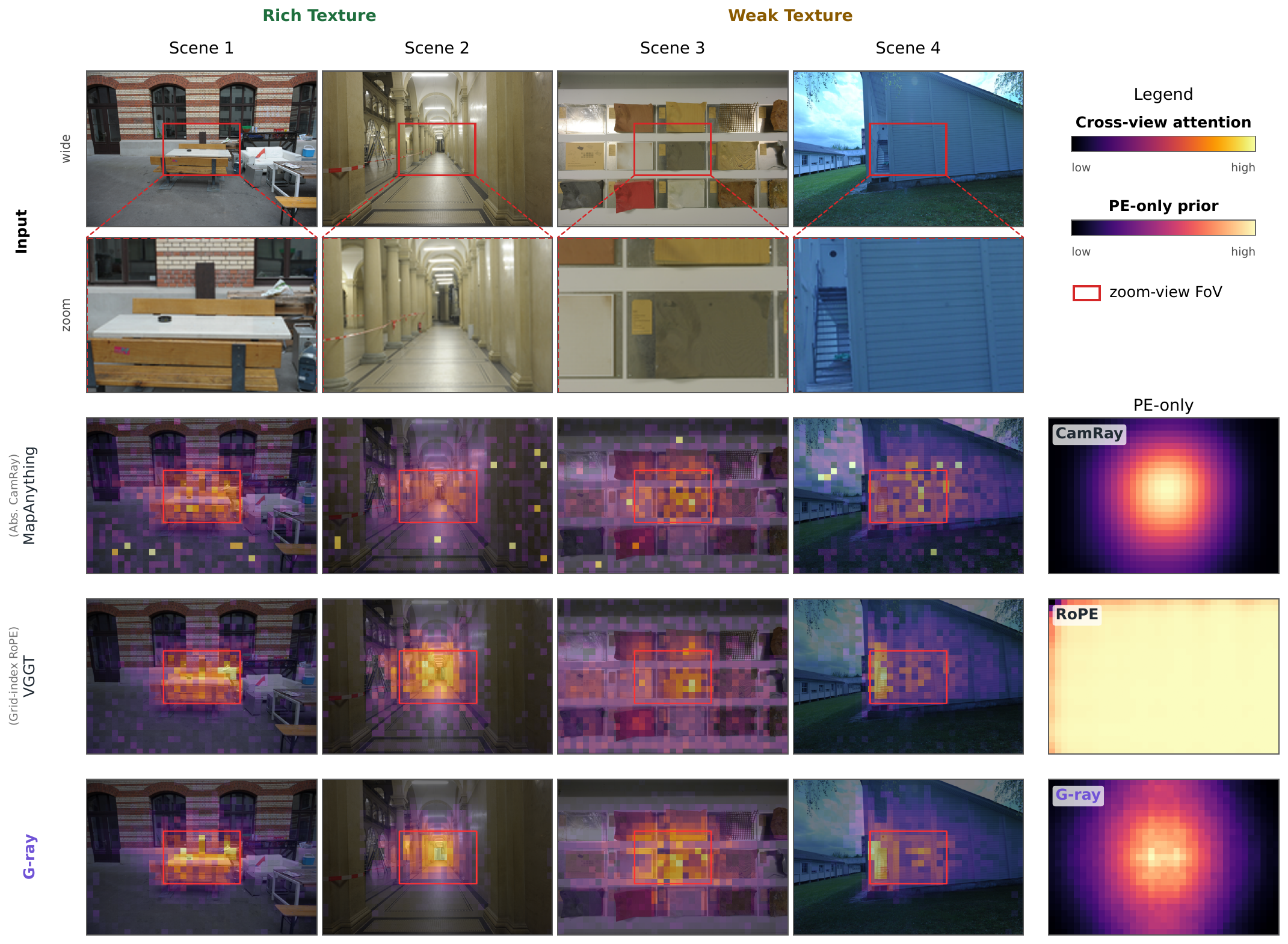}
  \caption{Cross-view attention under controlled FoV variation on ETH3D. The top two rows show wide-FoV images and their resized narrow-FoV crops. Red rectangles mark the shared FoV, and dashed lines link each region to its crop. Attention maps average wide-to-narrow responses over global-attention blocks and heads. The right column shows position-only priors (PE-only), computed for the rotary encodings using constant query and key features before rotation. In the weakly textured examples, \method{} concentrates attention more clearly within the shared FoV than grid-index RoPE. Its PE-only prior reflects angular alignment, whereas the grid-index prior is nearly uniform.}
  \label{fig:attention}
\end{figure*}

Fig.~\ref{fig:attention} illustrates this positional consistency on ETH3D image pairs under controlled FoV variation.
Each narrow-FoV view is a cropped and resized version of its wide-FoV counterpart.
Wide-view patch tokens serve as queries and narrow-view patch tokens as keys.
The maps average attention weights over global-attention blocks and heads and display the wide-view responses, with a red rectangle marking the shared FoV.
In the richly textured examples, all compared methods concentrate attention within the shared region.
In the weakly textured examples, grid-index RoPE assigns more attention outside that region, whereas \method{} maintains a clearer concentration within it.

The rightmost column shows position-only (PE-only) priors to separate the positional contribution from image content.
For rotary encodings, this probe assigns identical constant features to queries and keys before rotation.
The relative-phase identity for a bank of planar rotations parameterized by $\boldsymbol{\xi}$ is
\begin{equation}
\big(\mathbf{E}(\boldsymbol{\xi}_{i})\mathbf{q}\big)^{\!\top}
\big(\mathbf{E}(\boldsymbol{\xi}_{j})\mathbf{k}\big)
=
\mathbf{q}^{\top}
\mathbf{E}(\boldsymbol{\xi}_{j}-\boldsymbol{\xi}_{i})
\mathbf{k}
\label{eq:relative_phase}
\end{equation}
so the content-free interaction depends only on the coordinate difference $\Delta\boldsymbol{\xi}=\boldsymbol{\xi}_{j}-\boldsymbol{\xi}_{i}$.
With constant features, it reduces to a sum of cosines of frequency-weighted coordinate differences (Appendix~\ref{app:pe_only_score}).
For grid-index RoPE, these differences are measured in token-grid indices, and the displayed prior is nearly uniform.
For \method{}, $\boldsymbol{\xi}=(\theta_x,\theta_y)$, so the prior instead reflects camera-local ray-angle differences and favors small angular offsets in the displayed examples.
The multi-frequency cosine interaction need not decrease monotonically with angular separation.
These examples illustrate how ray-angle coordinates provide a positional prior aligned with the shared FoV, while the learned query and key features also shape the final attention distribution.

\subsubsection{Resolution-Independent Coordinate Range}

The ray-angle coordinate range does not grow with input resolution.
At a fixed FoV, changing the resolution alters the sampling density of the angular domain but leaves its extent unchanged.
After the image-to-ray map is updated, nearby token centers across resolutions correspond to nearby viewing rays and therefore similar ray-angle coordinates.
Exactly corresponding continuous image locations retain identical rays, as shown in Appendix~\ref{app:resampling_consistency}.
Token-grid indices, in contrast, span a larger range as resolution increases, requiring positional values outside the training range, a difficulty also studied in image classification~\cite{fan2024vitar}.
\method{} avoids this coordinate-range extrapolation through ray-level angular coordinates, since a higher-resolution input samples the same angular interval more finely.
Sec.~\ref{subsec:resolution_extrap} evaluates the complete network at up to twice the training resolution.

\subsubsection{Calibration Requirement}

These properties require view-specific image-to-ray calibration.
The angular map and the rotations are deterministic, so the encoding introduces no learned parameters and needs neither token depth nor camera extrinsics, although a host may keep such cues for its own purposes.
The image-to-ray map $\mathcal{U}_c$ is the interface to view-specific imaging geometry.
It can be obtained from pinhole intrinsics, a calibrated non-pinhole model such as \eqref{eq:fisheye}, or an external ray-map estimator.
The positional phases therefore depend on the accuracy of the supplied rays, regardless of how the map is obtained.
We therefore quantify controlled calibration error in Sec.~\ref{subsec:calibration_sensitivity} and off-the-shelf estimated image-to-ray geometry in Sec.~\ref{subsec:estimated_k}.

\subsection{Coordinate Replacement in Existing Encodings}
\label{subsec:replacement}

\method{} replaces the positional coordinates supplied to a host encoding's rotary interface.
Fig.~\ref{fig:replacement} summarizes the token-grid and projection-based cases.
Grid-index RoPE uses a patch index pair $(m,n)$.
\method{} instead maps the corresponding patch center through $\mathcal{U}_c$ and Eqs.~\eqref{eq:thetax} and~\eqref{eq:thetay} to obtain $(\theta_x,\theta_y)$ for the same rotary interface.
A projection-based encoding obtains a pixel pair $(u_q,v_q)$ by projecting a geometric quantity into the query camera.
\method{} converts this pair to query-camera ray angles while retaining the host's other geometric components.
When a host attaches several projected samples to one token, the same image-to-ray and ray-to-angle mapping is applied to each sample before the host's own positional aggregation.
Both paths therefore replace the image-plane positional coordinate rather than the geometric formulation built around it.

We evaluate one direct instantiation and two integrations with existing geometric encodings.
In the 3D reconstruction model, \method{} serves directly as the rotary position encoding in place of grid-index RoPE, allocating equal channel subspaces to $\theta_x$ and $\theta_y$.
The model also applies the value and output transformations in \eqref{eq:output}, enabling matched comparisons with grid-index RoPE and absolute CamRay conditioning.

In LVSM~\cite{jin2024lvsm} with GTA~\cite{miyato2023gta}, head dimensions are split between camera-transform and image-plane grid subspaces.
\method{} replaces only the grid coordinates, retaining the camera-transform subspace and CamRay conditioning.

In LVSM with the released RayRoPE implementation~\cite{wu2026rayrope}, \method{} replaces query-camera projected directional coordinates with ray angles.
The paired configurations retain camera-center components, query-frame depth, depth/interval predictors, and extrinsic projection.
Expected rotary coefficients are computed over uniform angular intervals defined by the transformed endpoints.
Appendix~\ref{app:freq_params} gives the explicit rotation blocks and frequency settings for these three host encodings.

\section{Experiments}
\label{sec:experiments}

We evaluate \method{} along four complementary dimensions.
Sec.~\ref{subsec:recon} compares multi-view 3D reconstruction under camera heterogeneity and homogeneous pinhole 3D reconstruction protocols.
Sec.~\ref{subsec:nvs} evaluates transfer to novel view synthesis through coordinate replacement in two host encodings.
Sec.~\ref{subsec:ablations} examines individual design choices under matched training conditions.
Sec.~\ref{subsec:analysis} examines robustness to controlled FoV variation and resolution changes, calibration requirements, and computational cost.

\subsection{Experimental Setup}
\label{subsec:setup}

\paragraph{Datasets}
We evaluate 3D reconstruction on three datasets spanning complementary forms of camera heterogeneity.
\textbf{PandaSet}~\cite{xiao2021pandaset} provides a rigid six-camera automotive rig with camera-specific intrinsics and FoVs.
On \textbf{ETH3D}~\cite{schops2017eth3d}, we evaluate controlled FoV variation, synthesized by center cropping and resizing with zoom factors $z \sim U[1.0,3.0]$ while retaining one native-FoV input as a wide-view anchor.
\textbf{ScanNet++}~\cite{yeshwanth2023scannetpp} is evaluated under three protocols with distinct roles.
The \emph{mixed-projection protocol} provides the main 3D reconstruction comparison, combining approximately equal numbers of DSLR frames under the calibrated four-coefficient fisheye model and undistorted pinhole iPhone frames from the same scene.
Covisibility-based sampling produces input sets of 2 to 50 views with both projection-model and device differences.
The \emph{undistorted mixed-device protocol} uses both streams after pinhole undistortion for the matched positional-encoding ablation and calibration diagnostics.
The \emph{high-resolution DSLR-only protocol} uses undistorted pinhole DSLR frames for the controlled FoV variation stress test in Fig.~\ref{fig:zoom}.

For NVS, \textbf{RealEstate10K}~\cite{zhou2018realestate10k} (RE10K) provides real-world videos for comparing a matched-FoV baseline with controlled FoV variation synthesized by center cropping and resizing.
One context view retains its native FoV, while the other context and each target independently sample zoom factors from $U[1,z_{\max}]$, with $z_{\max}\in\{1.5,2.0,2.5\}$.
The baseline $z{=}1.0$ applies no additional FoV change.
For evaluation, we preserve the 16:9 aspect ratio at $256{\times}144$ resolution with patch size 8, rather than using the $256{\times}256$ square training format.
This choice avoids upsampling after cropping typical $640{\times}360$ source frames up to $z{=}2.5$.
A $256{\times}256$ square evaluation would instead require upsampling beyond $z{\approx}1.4$.

\textbf{Objaverse}~\cite{deitke2023objaverse} evaluates NVS under camera heterogeneity, with the FoV independently sampled for each original view during rendering.
Two such views serve as context, and three regimes vary the target camera pose and FoV.
\emph{FoV change} keeps a context camera pose and applies an additional FoV change.
\emph{Novel viewpoint} places the target at a pose not seen in the context views, retaining its independently sampled FoV without further modification.
\emph{Novel viewpoint + FoV change} combines an unseen target pose with an additional change to its sampled FoV.
The two FoV-change regimes thus introduce further FoV variation beyond the differences already present in the independently sampled views.
Appendix~\ref{app:objaverse_protocols} defines the complete target sets, including protocols involving radial camera motion.

We also evaluate \emph{homogeneous pinhole 3D reconstruction protocols}~\cite{wang2025pi3} to assess compatibility when input views share one imaging geometry.
These protocols assess relative pose on TUM RGB-D~\cite{sturm2012tum} and Sintel~\cite{butler2012sintel}, video depth on Sintel and KITTI~\cite{geiger2012kitti}, and multi-view 3D reconstruction on NRGBD~\cite{azinovic2022neuralrgbd}, 7-Scenes~\cite{shotton2013scenes}, DTU~\cite{jensen2014dtu}, and undistorted ETH3D~\cite{schops2017eth3d}.
The undistorted ETH3D split in these protocols is distinct from the ETH3D setting with controlled FoV variation in Table~\ref{tab:recon_main}.

\paragraph{Metrics}
For the main 3D reconstruction evaluation with 2 to 50 input views, we report pointmap relative error (Points~rel~$\downarrow$), ray-direction error (Rays~err$^\circ$~$\downarrow$), depth relative error (Depth~rel~$\downarrow$), point norm-ratio inliers at $1.03$ (Pts~Inl.~$\uparrow$), pose ATE ($\downarrow$), and depth-ratio inliers at $1.03$ (Dp.~Inl.~$\uparrow$).
On PandaSet and ETH3D, Depth~rel and depth inliers use standard $z$-depth errors.
On mixed-projection ScanNet++, both non-pinhole DSLR and pinhole iPhone views are evaluated using \emph{ray depth}, the distance along the viewing ray.
The Depth~rel and Dp.~Inl.\ columns therefore use the depth convention appropriate to each dataset.
Inlier rates are reported as percentages at the dimensionless ratio threshold $1.03$; Appendix~\ref{app:metric_definitions} defines the point and depth tests.
Calibration diagnostics use 16 views.
For NVS, we report PSNR (dB~$\uparrow$), SSIM ($\uparrow$), and LPIPS ($\downarrow$), with aggregation defined in Appendix~\ref{app:nvs_aggregation}.
For the homogeneous pinhole 3D reconstruction protocols, we report trajectory ATE and RPE, video-depth AbsRel and $\delta{<}1.25$, and 3D reconstruction accuracy, completeness, and normal consistency.
3D reconstruction accuracy and completeness are reported in millimetres on DTU and in metres on the other datasets, without averaging across these units.

\paragraph{Implementation}
For 3D reconstruction, we use a VGGT-based architecture~\cite{wang2025vggt} with a DINOv2 ViT-L/14 backbone and alternating frame and global attention.
\method{} replaces grid-index RoPE in the attention aggregator and applies the value and output rotations in \eqref{eq:output}.
Training uses pinhole inputs from mixed indoor and outdoor datasets, first at a 224-pixel long side and then at 518 pixels, on 8 NVIDIA A100 GPUs.

All NVS variants share the LVSM decoder-only backbone~\cite{jin2024lvsm} and use 2 context views and 1 target view at $256{\times}256$ training resolution.
Each encoding variant is trained independently for 80\,k steps, separately on RE10K and Objaverse.
Appendix~\ref{app:training_details} provides the architecture details, training datasets, losses, and optimization schedules for both tasks.

\paragraph{Baselines}
3D reconstruction methods are grouped by whether external image-to-ray calibration is supplied.
With supplied calibration, we report \method{}, $\pi^3$-X~\cite{wang2025pi3}, WorldMirror, and MapAnything~\cite{keetha2025mapanything}.
For settings without externally supplied calibration, we report image-only DA3~\cite{lin2026da3}, DA3-Nested, MapAnything, $\pi^3$~\cite{wang2025pi3}, VGGT~\cite{wang2025vggt}, VGGT-$\Omega$~\cite{wang2026vggtomega}, OmniVGGT, and WorldMirror, together with \method{} driven by AnyCalib-estimated ray maps~\cite{tiradogarin2025anycalib}.
For non-pinhole projections, \method{}, MapAnything, and $\pi^3$-X receive calibrated ray maps; the evaluated WorldMirror interface accepts pinhole intrinsics only.
All methods in the heterogeneous 3D reconstruction comparison use the same sampled views and evaluation metrics, with input resizing adapted to each model's resolution requirements.

NVS baselines include Pl\"ucker-ray features with grid-index RoPE, GTA~\cite{miyato2023gta}, PRoPE~\cite{li2025prope}, and RayRoPE~\cite{wu2026rayrope}.
GTA + \method{} replaces token-grid position indices with camera-local ray-angle coordinates, while RayRoPE + \method{} replaces query-camera projected pixel coordinates.
Both variants retain the complementary geometric components of their host encoding.

\subsection{Multi-View 3D Reconstruction}
\label{subsec:recon}

\begin{table*}[!t]
\centering
\small
\setlength{\tabcolsep}{2.5pt}
\caption{Mean 50-view 3D reconstruction results across mixed-projection ScanNet++, PandaSet, and ETH3D under controlled FoV variation. The w/C group receives external calibration through each model's supported interface; WorldMirror uses pinhole intrinsics, while the other three methods receive calibrated ray maps. The w/o C group receives no external calibration and includes image-only methods and \method{} with AnyCalib-estimated ray maps. Bold and underlined values indicate the best and second-best results in each column.}
\label{tab:recon_main}
\begin{tabular}{lll cccccc}
\toprule
Setting & Method & Source / year & Points rel $\downarrow$ & Rays err$^\circ$ $\downarrow$ & Depth rel $\downarrow$ & Pts Inl.\ $\uparrow$ & Pose ATE $\downarrow$ & Dp.\ Inl.\ $\uparrow$ \\
\midrule
\multirow{4}{*}{w/C}
& \method{} & This work & \textbf{0.0744} & \textbf{0.38} & \textbf{0.0744} & \textbf{58.1} & \textbf{0.0221} & \textbf{51.9} \\
& $\pi^3$-X~\cite{wang2025pi3} & Release 2025$^{\dagger}$ & \underline{0.1130} & 2.32 & \underline{0.0798} & \underline{48.8} & \underline{0.0425} & \underline{42.3} \\
& WorldMirror~\cite{liu2026worldmirror} & ICML 2026 & 0.1338 & 2.26 & 0.0936 & 40.3 & 0.0441 & 38.2 \\
& MapAnything~\cite{keetha2025mapanything} & 3DV 2026 & 0.1372 & \underline{2.11} & 0.1199 & 37.5 & 0.0616 & 28.6 \\
\midrule
\multirow{9}{*}{w/o C}
& \method{} + AnyCalib & This work & 0.1347 & 2.49 & 0.1407 & 43.4 & 0.0702 & 24.4 \\
& DA3~\cite{lin2026da3} & ICLR 2026 & 0.3441 & 6.74 & 0.2827 & 25.3 & 0.1885 & 16.9 \\
& DA3-Nested~\cite{lin2026da3} & ICLR 2026$^{\ddagger}$ & 0.3278 & 6.04 & 0.2387 & 25.7 & 0.1444 & 15.8 \\
& MapAnything~\cite{keetha2025mapanything} & 3DV 2026 & 0.2404 & 3.90 & 0.2037 & 27.0 & 0.1208 & 14.0 \\
& $\pi^3$~\cite{wang2025pi3} & ICLR 2026 & 0.2828 & 4.29 & 0.1939 & 34.8 & 0.1537 & 21.3 \\
& VGGT~\cite{wang2025vggt} & CVPR 2025 & 0.3359 & 4.31 & 0.2081 & 22.8 & 0.1795 & 19.4 \\
& VGGT-$\Omega$~\cite{wang2026vggtomega} & CVPR 2026 & 0.2011 & 4.12 & 0.1329 & 46.1 & 0.0780 & 38.1 \\
& OmniVGGT~\cite{peng2025omnivggt} & CVPR 2026 & 0.2708 & 4.30 & 0.3198 & 25.8 & 0.1815 & 13.9 \\
& WorldMirror~\cite{liu2026worldmirror} & ICML 2026 & 0.1944 & 3.34 & 0.1554 & 32.0 & 0.0839 & 25.1 \\
\bottomrule
\end{tabular}
\par\smallskip
\begin{minipage}{\textwidth}
\footnotesize
$^{\dagger}$\,$\pi^3$-X is the authors' \href{https://github.com/yyfz/Pi3}{\textcolor{blue}{December 2025 model release}}, not a separate conference publication.
$^{\ddagger}$\,DA3-Nested is a released DA3 variant and shares its publication.
\end{minipage}
\end{table*}

\begin{figure*}[!t]
  \centering
  \includegraphics[width=\textwidth]{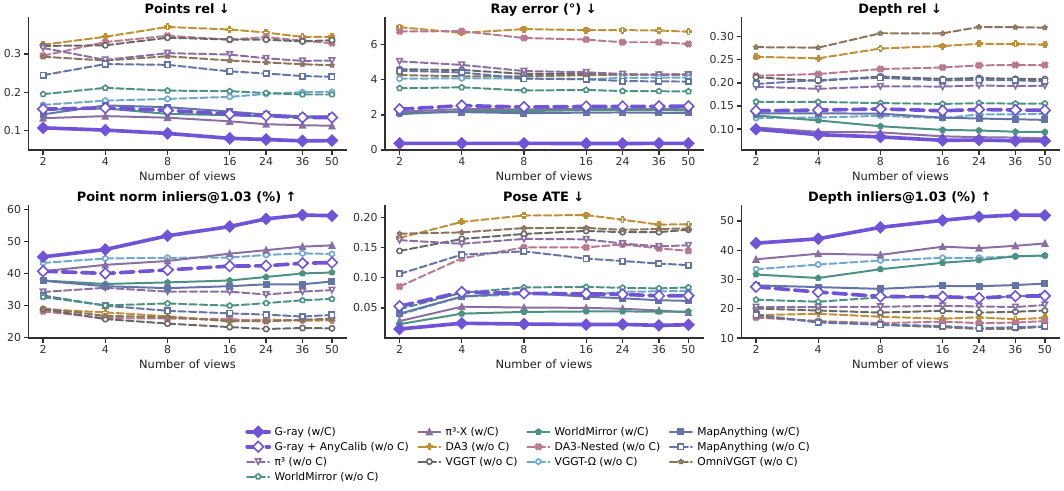}
  \caption{Mean 3D reconstruction performance across mixed-projection ScanNet++, PandaSet, and ETH3D with 2 to 50 input views. ScanNet++ uses ray-depth metrics. The dashed \method{} curves use AnyCalib-estimated ray maps. Table~\ref{tab:recon_main} gives the 50-view results. Lower is better for pointmap, ray-direction, depth, and pose errors. Higher is better for point and depth inlier rates.}
  \label{fig:recon_views}
\end{figure*}

\begin{figure*}[!t]
  \centering
  \includegraphics[width=\textwidth]{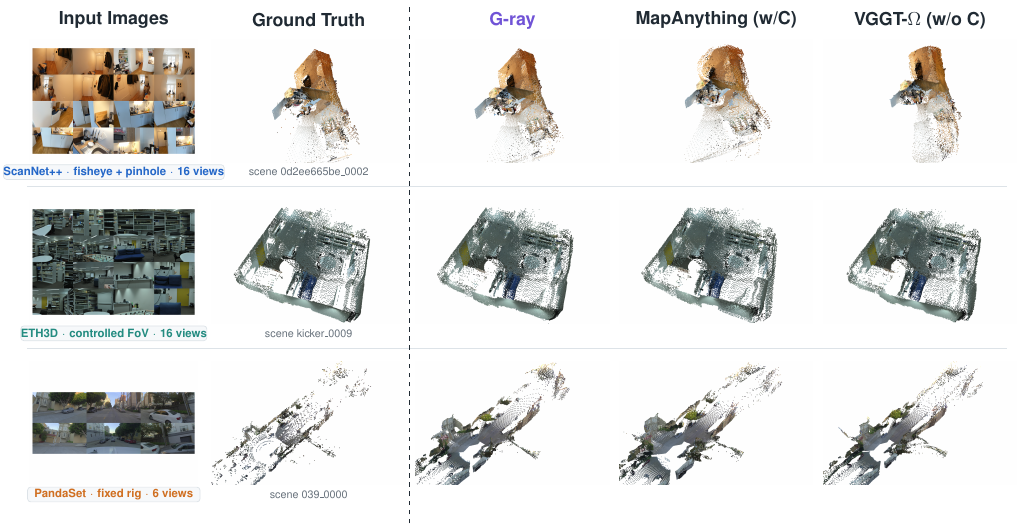}
  \caption{Qualitative 3D reconstruction under camera heterogeneity. Rows show mixed-projection ScanNet++, ETH3D under controlled FoV variation, and the PandaSet multi-camera rig. Columns show inputs, ground truth, \method{}, MapAnything with supplied calibration, and image-only VGGT-$\Omega$. Within each row, all point clouds share ground-truth-defined display bounds and identical rendering viewpoints, FoVs, and settings. PandaSet ground truth is sparse LiDAR, whereas the predictions are dense image-based 3D reconstructions.}
  \label{fig:recon_qualitative}
\end{figure*}

We first evaluate 3D reconstruction under camera heterogeneity, then examine compatibility with homogeneous pinhole inputs.
Table~\ref{tab:recon_main} reports three-dataset averages at 50 views, Fig.~\ref{fig:recon_views} summarizes performance from 2 to 50 views, and Fig.~\ref{fig:recon_qualitative} compares reconstructed scene geometry.
Complete per-dataset results are provided in Appendix Table~\ref{tab:recon_per_dataset}.

\subsubsection{Heterogeneous Protocols}
\label{subsec:recon_hetero}

With supplied calibration, \method{} leads all six cross-dataset metrics at 50 views, including 0.0744 pointmap relative error and 58.1\% point inliers (Table~\ref{tab:recon_main}).
Without externally supplied calibration, \method{} + AnyCalib leads on average pointmap error, ray-direction error, and pose ATE, while VGGT-$\Omega$ leads on depth error and point and depth inliers.

The clearest gains occur on mixed-projection ScanNet++, where the pinhole-trained model processes both pinhole and non-pinhole inputs through their ray maps without retraining.
With ground-truth ray maps, \method{} leads all six metrics, achieving 0.0583 pointmap relative error and 75.1\% point inliers.
On PandaSet and ETH3D, \method{} achieves the lowest ray-direction errors ($0.27^\circ$ and $0.15^\circ$), while $\pi^3$-X and VGGT-$\Omega$, respectively, lead most other metrics.
The gains therefore depend on the form of camera heterogeneity rather than extending uniformly to every benchmark and metric.
Appendix~\ref{app:recon_per_dataset} gives the complete per-dataset results and analysis.

\subsubsection{Homogeneous Pinhole 3D Reconstruction Protocols}
\label{subsec:pi3_protocol}

The preceding benchmarks contain cross-view differences in FoV or projection model.
We next examine performance when input views share a homogeneous pinhole imaging geometry.
Table~\ref{tab:pi3_homogeneous} summarizes relative pose, video depth, and multi-view 3D reconstruction under the homogeneous pinhole 3D reconstruction protocols~\cite{wang2025pi3}.
Complete results appear in Appendix~\ref{app:pi3_protocol}.

VGGT-$\Omega$ achieves the strongest overall results under these protocols.
\method{} with supplied calibration attains the second-best Sintel pose ATE (0.110 vs.\ 0.171 for VGGT and 0.219 for MapAnything) and the second-best indoor 3D reconstruction accuracy on NRGBD and 7-Scenes, including 0.019 Acc on dense 7-Scenes versus 0.018 for VGGT-$\Omega$.
It also improves over MapAnything with the same calibrated input on TUM pose (0.0122 vs.\ 0.0179 ATE) and on all four indoor Acc columns.
It trails VGGT-$\Omega$, and on Sintel video depth, KITTI AbsRel, DTU Acc, and undistorted ETH3D Acc it also trails VGGT.

The 3D reconstruction model with \method{} retains competitive accuracy under homogeneous pinhole inputs, supporting the compatibility of ray-level angular encoding when camera heterogeneity is absent.

\begin{table*}[tp]
\centering
\small
\setlength{\tabcolsep}{3.2pt}
\caption{Selected results under homogeneous pinhole 3D reconstruction protocols~\cite{wang2025pi3}. \method{} and MapAnything receive ground-truth pinhole calibration, while VGGT and VGGT-$\Omega$ are image-only. Indoor 3D reconstruction accuracy (Acc) is in metres. Bold and underlined values indicate the best and second-best results in each column. Appendix~\ref{app:pi3_protocol} provides the complete results.}
\label{tab:pi3_homogeneous}
\begin{tabular}{ll cc cc cccc}
\toprule
Method & Source / year & \multicolumn{2}{c}{Rel.\ pose ATE $\downarrow$} & \multicolumn{2}{c}{Video depth AbsRel $\downarrow$} & \multicolumn{4}{c}{Indoor Acc $\downarrow$} \\
\cmidrule(lr){3-4}\cmidrule(lr){5-6}\cmidrule(lr){7-10}
 & & TUM & Sintel & Sintel & KITTI & NRGBD-s & NRGBD-d & 7Sc-s & 7Sc-d \\
\midrule
\method{} & This work & 0.0122 & \underline{0.110} & 0.357 & 0.092 & \underline{0.037} & \underline{0.023} & \underline{0.038} & \underline{0.019} \\
MapAnything~\cite{keetha2025mapanything} & 3DV 2026 & 0.0179 & 0.219 & 0.401 & 0.083 & 0.088 & 0.036 & 0.059 & 0.023 \\
VGGT~\cite{wang2025vggt} & CVPR 2025 & \underline{0.0090} & 0.171 & \underline{0.324} & \underline{0.082} & 0.058 & 0.029 & 0.055 & 0.025 \\
VGGT-$\Omega$~\cite{wang2026vggtomega} & CVPR 2026 & \textbf{0.0044} & \textbf{0.040} & \textbf{0.251} & \textbf{0.072} & \textbf{0.022} & \textbf{0.014} & \textbf{0.029} & \textbf{0.018} \\
\bottomrule
\end{tabular}
\end{table*}

\subsection{Coordinate Replacement in Novel View Synthesis}
\label{subsec:nvs}

\begin{table*}[!t]
\centering
\caption{RE10K NVS at $256{\times}144$ under controlled FoV variation. No extra FoV change uses $z{=}1$; otherwise, one context stays at its native FoV and other views sample $z\sim U[1,z_{\max}]$. All methods use the same 6,202 test scenes. PSNR is in dB; SSIM and LPIPS complete the quality comparison. Bold indicates the best unrounded value per column. Appendix~\ref{app:nvs_aggregation} gives the metric definitions.}
\label{tab:nvs_re10k}
\footnotesize
\setlength{\tabcolsep}{3pt}
\begin{tabular*}{\textwidth}{@{\extracolsep{\fill}}l rrr rrr rrr rrr @{}}
\toprule
Method & \multicolumn{3}{c}{No extra FoV change} & \multicolumn{3}{c}{$z_{\max}{=}1.5$} & \multicolumn{3}{c}{$z_{\max}{=}2.0$} & \multicolumn{3}{c}{$z_{\max}{=}2.5$} \\
\cmidrule(lr){2-4}\cmidrule(lr){5-7}\cmidrule(lr){8-10}\cmidrule(lr){11-13}
 & PSNR$\uparrow$ & SSIM$\uparrow$ & LPIPS$\downarrow$ & PSNR$\uparrow$ & SSIM$\uparrow$ & LPIPS$\downarrow$ & PSNR$\uparrow$ & SSIM$\uparrow$ & LPIPS$\downarrow$ & PSNR$\uparrow$ & SSIM$\uparrow$ & LPIPS$\downarrow$ \\
\midrule
Pl\"ucker~\cite{jin2024lvsm} & 20.67 & 0.644 & 0.191 & 18.21 & 0.523 & 0.233 & 17.50 & 0.493 & 0.276 & 17.05 & 0.482 & 0.313 \\
PRoPE~\cite{li2025prope} & 19.80 & 0.625 & 0.202 & 15.18 & 0.416 & 0.331 & 13.80 & 0.376 & 0.422 & 13.03 & 0.357 & 0.478 \\
\midrule
GTA~\cite{miyato2023gta} & 19.06 & 0.595 & 0.213 & 15.11 & 0.415 & 0.342 & 14.00 & 0.388 & 0.421 & 13.44 & 0.380 & 0.463 \\
GTA + \textbf{\method{}} & 20.62 & 0.651 & 0.178 & 20.72 & 0.632 & 0.181 & 20.41 & 0.608 & 0.204 & 20.12 & 0.588 & 0.230 \\
\midrule
RayRoPE~\cite{wu2026rayrope} & 21.35 & 0.692 & 0.152 & 20.98 & 0.651 & 0.177 & 20.65 & 0.624 & 0.208 & 20.32 & 0.605 & 0.242 \\
RayRoPE + \textbf{\method{}} & \textbf{21.47} & \textbf{0.706} & \textbf{0.142} & \textbf{22.09} & \textbf{0.719} & \textbf{0.137} & \textbf{21.78} & \textbf{0.703} & \textbf{0.155} & \textbf{21.42} & \textbf{0.685} & \textbf{0.179} \\
\bottomrule
\end{tabular*}
\end{table*}

\begin{table*}[!t]
\centering
\caption{Objaverse NVS on 8\,000 objects at $256{\times}256$. Context FoVs are independently sampled. FoV change keeps a context pose; Novel viewpoint uses an unseen pose at its sampled FoV; the last regime adds FoV change at an unseen pose. PSNR is in dB. Bold indicates the best unrounded value. Appendix~\ref{app:nvs_aggregation} defines the target-wise and object-wise aggregation.}
\label{tab:nvs_splits}
\footnotesize
\setlength{\tabcolsep}{3pt}
\begin{tabular*}{\textwidth}{@{\extracolsep{\fill}}l rrr rrr rrr @{}}
\toprule
Method & \multicolumn{3}{c}{FoV change} & \multicolumn{3}{c}{Novel viewpoint} & \multicolumn{3}{c}{Novel viewpoint + FoV change} \\
\cmidrule(lr){2-4}\cmidrule(lr){5-7}\cmidrule(lr){8-10}
 & PSNR$\uparrow$ & SSIM$\uparrow$ & LPIPS$\downarrow$ & PSNR$\uparrow$ & SSIM$\uparrow$ & LPIPS$\downarrow$ & PSNR$\uparrow$ & SSIM$\uparrow$ & LPIPS$\downarrow$ \\
\midrule
Pl\"ucker~\cite{jin2024lvsm} & 26.20 & 0.892 & 0.129 & 18.88 & 0.863 & 0.197 & 11.17 & 0.740 & 0.400 \\
PRoPE~\cite{li2025prope} & 26.07 & 0.890 & 0.120 & 19.68 & 0.868 & 0.177 & 10.67 & 0.734 & 0.413 \\
\midrule
GTA~\cite{miyato2023gta} & 25.72 & 0.886 & 0.128 & 19.11 & 0.864 & 0.191 & 10.92 & 0.740 & 0.402 \\
GTA + \textbf{\method{}} & 26.50 & 0.895 & 0.108 & 19.55 & 0.867 & 0.182 & 11.80 & 0.752 & 0.370 \\
\midrule
RayRoPE~\cite{wu2026rayrope} & \textbf{28.42} & \textbf{0.919} & \textbf{0.073} & 19.55 & 0.868 & 0.180 & 10.68 & 0.734 & 0.417 \\
RayRoPE + \textbf{\method{}} & 27.97 & 0.912 & 0.083 & \textbf{20.03} & \textbf{0.873} & \textbf{0.164} & \textbf{12.00} & \textbf{0.764} & \textbf{0.345} \\
\bottomrule
\end{tabular*}
\end{table*}

We evaluate whether ray-angle coordinate replacement improves NVS in both grid-index and projection-based rotary encodings.
Within each host, paired variants share the LVSM backbone, complementary geometric components, and training budget.
\method{} replaces the image-plane positional coordinates with ray-angle coordinates, with per-host frequency settings specified in Appendix~\ref{app:freq_params}.

\paragraph{RealEstate10K}
Ray-angle coordinates improve both host encodings under controlled FoV variation (Table~\ref{tab:nvs_re10k}).
At $z_{\max}{=}2.5$, GTA + \method{} achieves 20.12~dB, compared with 13.44~dB for GTA and 13.03~dB for PRoPE.
RayRoPE + \method{} obtains the highest PSNR and SSIM and the lowest LPIPS in every setting, with gains over RayRoPE of 1.11, 1.13, and 1.10~dB at $z_{\max}{=}1.5$, 2.0, and 2.5, respectively.
The corresponding gain on the matched-FoV baseline is 0.12~dB.
These results support the use of ray-angle coordinates when synthesizing real-world scenes from views with different imaging geometries.

\paragraph{Objaverse}
Both host encodings benefit from \method{} when synthesizing unseen viewpoints under heterogeneous FoVs (Table~\ref{tab:nvs_splits}).
GTA + \method{} improves PSNR by 0.78~dB for FoV change, 0.44~dB for Novel viewpoint, and 0.88~dB for Novel viewpoint + FoV change.
The RayRoPE integration yields gains of 0.48~dB at novel viewpoints and 1.32~dB when their native FoVs are additionally changed, while retaining the depth/interval predictors and camera transformations.
SSIM and LPIPS also improve for both hosts at novel viewpoints, with and without additional FoV change.
These paired results support applying the same coordinate replacement to two distinct rotary formulations.

The benefit depends on the context-to-target relationship.
When the target shares a pose with one context view, RayRoPE + \method{} scores 27.97~dB versus 28.42~dB for RayRoPE, a decrease of 0.46~dB computed before rounding.
Thus, the benefit is not uniform across all FoV changes and is larger for synthesis at unseen viewpoints.
Additional results involving radial camera motion are reported in Appendix~\ref{app:objaverse_results}.

\begin{figure*}[!t]
  \centering
  \includegraphics[width=\textwidth]{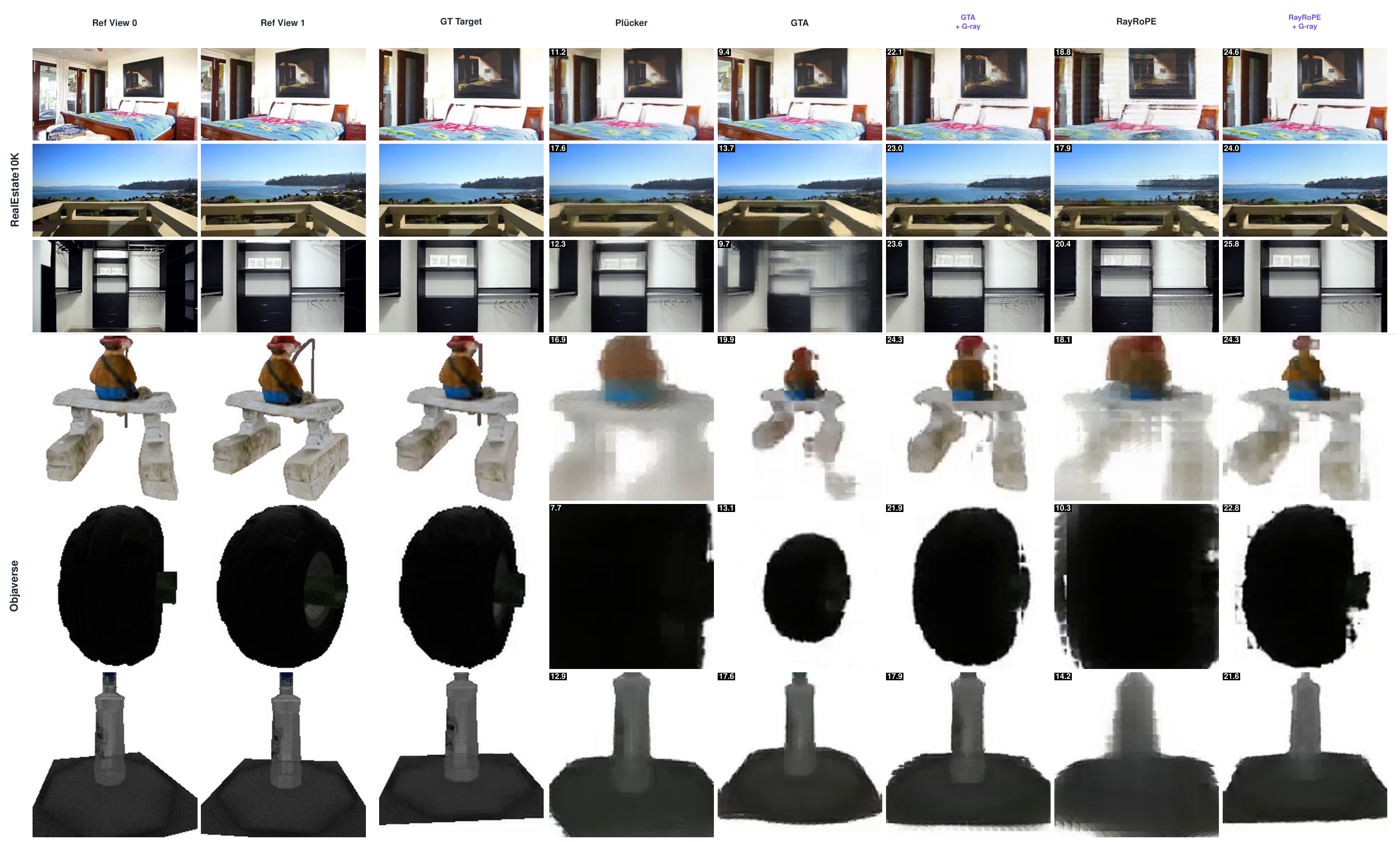}
  \caption{Qualitative NVS under FoV heterogeneity.
  The upper block shows RE10K under controlled FoV variation at $256{\times}144$ resolution and $z_{\max}{=}2.0$. One context retains its native FoV, while the remaining views sample zoom factors from $U[1,2]$.
  The lower block shows Objaverse with changing FoV along short viewpoint trajectories.
  Columns contain two context views, the ground-truth target, and predictions from Pl\"ucker, GTA, GTA + \method{}, RayRoPE, and RayRoPE + \method{}.
  Numbers report target-view PSNR (dB $\uparrow$). Both \method{} variants better preserve scene structure in these examples.}
  \label{fig:nvs_qual}
\end{figure*}

\paragraph{Qualitative Comparison}
Fig.~\ref{fig:nvs_qual} illustrates the effects of coordinate replacement under FoV variation.
In the RE10K examples, GTA and Pl\"ucker exhibit blur and structural errors, while GTA + \method{} better preserves scene layout.
RayRoPE + \method{} reduces the ghosting and banding visible in baseline RayRoPE predictions.
The Objaverse examples show similar improvements along short viewpoint trajectories with changing FoV.
These examples illustrate the structural improvements reflected in the quantitative comparisons.
Additional target-view comparisons are provided in Appendix~\ref{app:nvs_visualizations}.

\subsection{Ablation Studies}
\label{subsec:ablations}

The ablations examine coordinate representation, value and output rotations, and frequency scale under matched training conditions.

\subsubsection{Coordinate Replacement under a Matched 3D Reconstruction Protocol}
\label{subsec:ablation_recon}

To isolate the positional encoding, all variants in Table~\ref{tab:ablation_recon} use the same ViT-S/14 backbone and 45-epoch training schedule.

\begin{table}[!t]
\centering
\footnotesize
\setlength{\tabcolsep}{2.5pt}
\caption{Matched positional-encoding ablation for 3D reconstruction. All variants use ViT-S/14 with embedding dimension 384 and the same training schedule. ScanNet++ uses approximately equal numbers of undistorted pinhole DSLR and iPhone views. ETH3D uses controlled FoV variation. Bold values indicate the best result in each column.}
\label{tab:ablation_recon}
\begin{tabular}{l ccc ccc}
\toprule
 & \multicolumn{3}{c}{ScanNet++} & \multicolumn{3}{c}{ETH3D (FoV var.)} \\
\cmidrule(lr){2-4} \cmidrule(lr){5-7}
Variant & Pts\,$\downarrow$ & Ray$^\circ$\,$\downarrow$ & Pose\,$\downarrow$ & Pts\,$\downarrow$ & Ray$^\circ$\,$\downarrow$ & Pose\,$\downarrow$ \\
\midrule
\method{} (full) & \textbf{0.820} & \textbf{15.9} & \textbf{0.408} & \textbf{0.984} & \textbf{11.1} & \textbf{0.385} \\
\quad w/o $v,o$ encoding & 0.927 & 21.2 & 0.444 & 0.996 & 13.3 & 0.389 \\
\quad$\to$ Grid-index RoPE & 0.970 & 19.8 & 0.461 & 1.077 & 13.0 & 0.394 \\
\quad$\to$ Absolute CamRay & 0.945 & 19.8 & 0.438 & 1.068 & 13.2 & 0.393 \\
\bottomrule
\end{tabular}
\end{table}

\paragraph{Ray-Angle PE vs.\ Image-Plane Grid-Index RoPE}
\method{} outperforms grid-index RoPE on all six displayed entries across both datasets.
On ScanNet++, pointmap relative error, ray-direction error, and pose ATE decrease by approximately 15.5\%, 19.7\%, and 11.5\%, respectively.
On ETH3D under controlled FoV variation, the corresponding reductions are 8.6\%, 14.6\%, and 2.3\%.
The improvement supports the ray-angle encoding design over the evaluated grid-index baseline.

\paragraph{Ray-Angle PE vs.\ Absolute CamRay}
\method{} also outperforms absolute CamRay conditioning on all six columns.
Relative reductions in pointmap error, ray-direction error, and pose ATE are 13.2\%, 19.7\%, and 6.8\% on ScanNet++, and 7.9\%, 15.9\%, and 2.0\% on ETH3D.
Under this matched setting, relative ray-angle encoding is more effective than the evaluated absolute camera-ray conditioning design.

\paragraph{Value and Output Encoding}
Applying the rotations only to queries and keys, without the value and inverse output rotations, degrades all six displayed entries.
ScanNet++ ray error increases from $15.9^\circ$ to $21.2^\circ$ (+33\%) and ETH3D ray error from $11.1^\circ$ to $13.3^\circ$ (+20\%).
The result supports applying the rotations to values and outputs in addition to queries and keys.

\subsubsection{NVS Frequency Scale}
\label{subsec:ablation_nvs}

\begin{table}[!t]
\centering
\caption{Frequency-scale ablation for GTA\,+\,\method{} on RE10K, with frequency base fixed at $\beta{=}100$. PSNR (dB $\uparrow$) is measured at $256{\times}144$ resolution under the FoV protocols of Table~\ref{tab:nvs_re10k}. Bold values indicate the best result in each column. The selected setting is $\alpha{=}10$.}
\label{tab:ablation_fs}
\resizebox{\columnwidth}{!}{%
\begin{tabular}{c cccc}
\toprule
& Homogeneous & \multicolumn{3}{c}{Heterogeneous imaging geometry} \\
\cmidrule(lr){2-2}\cmidrule(lr){3-5}
$\alpha$ & $z{=}1.0$ & $z_{\max}{=}1.5$ & $z_{\max}{=}2.0$ & $z_{\max}{=}2.5$ \\
\midrule
1  & 20.66 & 20.40 & 19.98 & 19.65 \\
2  & \textbf{20.97} & \textbf{20.74} & 20.33 & 20.01 \\
4  & 20.77 & 20.63 & 20.26 & 19.98 \\
6  & 20.76 & 20.53 & 20.08 & 19.72 \\
8  & 20.54 & 20.58 & 20.30 & 20.04 \\
10 & 20.62 & 20.72 & \textbf{20.41} & \textbf{20.12} \\
15 & 20.31 & 20.50 & 20.23 & 19.96 \\
20 & 20.39 & 20.51 & 20.19 & 19.90 \\
40 & 20.25 & 20.60 & 20.31 & 20.01 \\
\bottomrule
\end{tabular}
}
\end{table}

Table~\ref{tab:ablation_fs} evaluates the frequency scale $\alpha$ of \eqref{eq:frequency} in GTA\,+\,\method{}.
The matched-FoV baseline and mild FoV variation ($z_{\max}{=}1.5$) achieve their highest PSNR at $\alpha{=}2$, with 20.97 and 20.74~dB, respectively.
Stronger FoV variation ($z_{\max}{=}2.0$ and 2.5) favors $\alpha{=}10$, with 20.41 and 20.12~dB.
This setting also gives the lowest LPIPS under all four protocols (Appendix~\ref{app:nvs_frequency}).
We use $\alpha{=}10$ consistently for GTA\,+\,\method{}, balancing performance under larger FoV differences with a 0.35~dB difference from the matched-FoV PSNR maximum.

\subsection{Analysis}
\label{subsec:analysis}

Following the attention diagnostic in Sec.~\ref{subsec:attn_diag}, we evaluate robustness to controlled FoV variation and resolution changes, dependence on calibration, and inference cost.

\subsubsection{Controlled FoV Variation}
\label{subsec:zoom}

\begin{figure}[!t]
  \centering
  \includegraphics[width=\columnwidth]{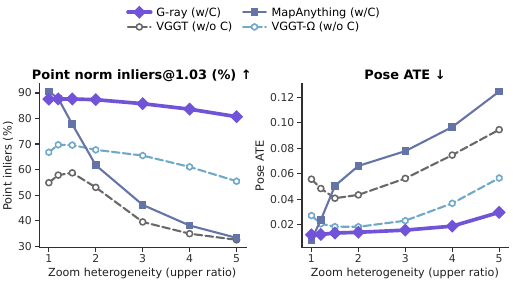}
  \caption{Controlled FoV variation on ScanNet++ using undistorted pinhole DSLR frames. One view retains its native FoV, while the remaining views are center-cropped and resized with zoom factors sampled uniformly up to $z_\text{high}\in\{1.0,\ldots,5.0\}$. The panels report point inlier rate (\%, $\uparrow$) and pose ATE ($\downarrow$). \method{} remains comparatively stable as cross-view FoV differences increase.}
  \label{fig:zoom}
\end{figure}

Fig.~\ref{fig:zoom} evaluates increasing FoV disparity on 15 high-resolution ScanNet++ scenes using undistorted pinhole DSLR frames.
For each 16-view input, one view retains its native FoV and the remaining views uniformly sample zoom factors up to $z_\text{high}\in\{1.0,\ldots,5.0\}$.
The source frames and camera poses remain fixed, while center cropping and resizing change the angular coverage of each view.
As $z_\text{high}$ increases, MapAnything with supplied calibration exhibits fewer point inliers and higher pose ATE, whereas \method{} remains comparatively stable on both metrics.
VGGT-$\Omega$ improves over VGGT but still trails \method{} at large FoV differences.
Together with the matched ablation in Table~\ref{tab:ablation_recon}, this result supports robustness to controlled FoV variation within the pinhole projection model.

\subsubsection{Resolution Extrapolation}
\label{subsec:resolution_extrap}

\begin{figure}[!t]
  \centering
  \includegraphics[width=\columnwidth]{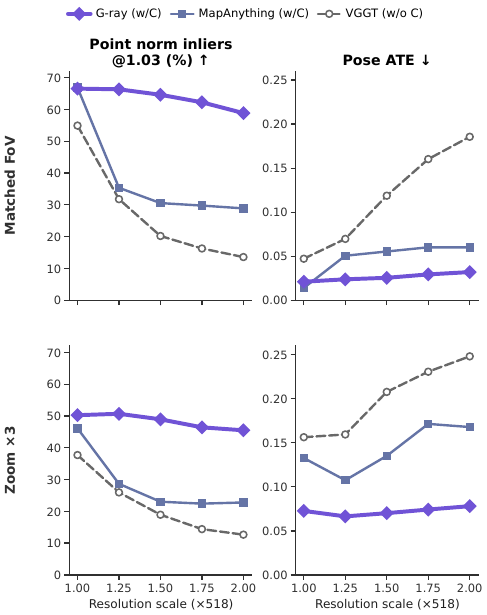}
  \caption{Resolution extrapolation on ETH3D. Columns report point inlier rate (\%, $\uparrow$) and pose ATE ($\downarrow$), with shared axis ranges across rows. The top row uses matched FoVs. The bottom row uses controlled FoV variation with one native-FoV anchor and uniformly sampled zoom factors up to $z_\text{high}{=}3$. Input long-side resolutions are patch-aligned nominal scales of $518\times\{1.0,1.25,1.5,1.75,2.0\}$. \method{} degrades more gradually than MapAnything with supplied calibration and image-only VGGT.}
  \label{fig:res_extrap}
\end{figure}

Although ray-angle coordinates do not explicitly depend on pixel scale, resolution generalization must be evaluated for the complete network.
We test input resolutions above the 518-pixel training long side.
On ETH3D, we sweep patch-aligned long-side resolutions corresponding to nominal scales $\{1.0,1.25,1.5,1.75,2.0\}\times518$ at 4 views, and compare \method{} against MapAnything with calibrated input and VGGT.
Fig.~\ref{fig:res_extrap} compares matched-FoV inputs with controlled FoV variation that retains one native-FoV view and samples zoom factors uniformly up to $z_\text{high}{=}3$.

Without FoV variation, doubling the input resolution reduces point inliers from 66.9\% to 28.9\% for MapAnything and from 54.9\% to 13.6\% for VGGT, compared with 66.5\% to 58.8\% for \method{}.
Under the $z_\text{high}{=}3$ protocol the absolute accuracy is lower for all methods, and the same ordering persists, with \method{} declining from 50.2\% to 45.5\% while MapAnything and VGGT again drop substantially.
Pose ATE follows the same pattern.
The results support improved resolution transfer on ETH3D with both matched and heterogeneous FoVs.

\subsubsection{Sensitivity to Intrinsic Calibration}
\label{subsec:calibration_sensitivity}

Because \method{} derives relative phases from view-specific calibration, we distinguish sensitivity to calibration error from robustness to FoV differences in correctly calibrated images.
We perturb only the conditioning intrinsics while leaving images, poses, ground-truth rays, and depth unchanged.
The focal trials multiply $f_x$ and $f_y$ by $1+\epsilon$ for $\epsilon \in \{0.05,0.10,0.20\}$ while fixing the principal point.
The shuffled condition cyclically assigns intrinsics from another view in the same input set.
Table~\ref{tab:intrinsic_noise} reports averages over ETH3D, ScanNet++, and PandaSet at 16 views.
ScanNet++ uses the undistorted mixed-device protocol.

\begin{table}[!t]
\centering
\caption{Sensitivity to intrinsic-calibration errors, averaged over ETH3D, undistorted mixed-device ScanNet++, and PandaSet at 16 views. Only the supplied intrinsics change, while images and evaluation targets remain fixed. Lower is better except for point inlier rate. Bold values indicate the best result in each column.}
\label{tab:intrinsic_noise}
\resizebox{\columnwidth}{!}{%
\begin{tabular}{l ccccc}
\toprule
Condition & Points rel & Rays err$^\circ$ & Depth rel & Pose ATE & Pts Inl. \\
\midrule
Correct $K$ & \textbf{0.0756} & \textbf{0.191} & \textbf{0.0733} & \textbf{0.0213} & \textbf{57.7} \\
$+5\%$ focal & 0.0881 & 0.995 & 0.0852 & 0.0280 & 51.8 \\
$+10\%$ focal & 0.1087 & 1.899 & 0.1035 & 0.0393 & 43.0 \\
$+20\%$ focal & 0.1540 & 3.552 & 0.1412 & 0.0638 & 31.3 \\
Shuffled $K$ & 0.4770 & 5.778 & 0.3880 & 0.2925 & 12.0 \\
\bottomrule
\end{tabular}%
}
\end{table}

All five reported metrics degrade monotonically as the uniform focal perturbation increases.
At $+5\%$, point and depth errors remain substantially closer to the correct-calibration condition than to the shuffled condition, but ray-direction error already rises from $0.191^\circ$ to $0.995^\circ$.
Assigning intrinsics to the wrong views produces the largest degradation on every metric.
Thus, robustness to cross-view FoV variation does not remove the need for accurate, correctly associated image-to-ray calibration.

\subsubsection{Off-the-Shelf Intrinsic Estimates}
\label{subsec:estimated_k}

To test \method{} without supplied calibration, we use the single-image estimator AnyCalib~\cite{tiradogarin2025anycalib} to recover image-to-ray geometry.
Table~\ref{tab:estimated_k}(a) reports calibration accuracy on three datasets, and panel (b) evaluates 16-view ScanNet++ 3D reconstruction with estimated conditioning geometry while retaining ground-truth rays and depth as evaluation targets.
For the undistorted mixed-device protocol, we also compare multi-view VGGT estimates~\cite{wang2025vggt}.

\begin{table}[!t]
\centering
\footnotesize
\setlength{\tabcolsep}{3pt}
\caption{Estimated image-to-ray geometry and 3D reconstruction accuracy. (a) AnyCalib calibration accuracy, evaluated independently of 3D reconstruction. (b) \method{} 3D reconstruction on ScanNet++ using estimated calibration, with ground-truth rays and depth retained as evaluation targets. ScanNet++ in (a) and Undist.\ mixed in (b) use undistorted pinhole DSLR and iPhone views. Mixed proj.\ combines non-pinhole DSLR and pinhole iPhone views. Panel (b) uses the abbreviated metrics of Table~\ref{tab:recon_main}. Bold values indicate the best result for each metric within each protocol, including ground-truth calibration.}
\label{tab:estimated_k}

\smallskip
(a) Calibration accuracy
\smallskip

\begin{tabular}{l ccc}
\toprule
Dataset & Rays err$^\circ$ $\downarrow$ & Rel. focal $\downarrow$ & PP err px $\downarrow$ \\
\midrule
ETH3D & 1.47 & 7.1\% & 1.44 \\
ScanNet++ & 2.04 & 8.6\% & 0.78 \\
PandaSet & 4.53 & 24.7\% & 22.35 \\
Average & 2.68 & 13.5\% & 8.19 \\
\bottomrule
\end{tabular}

\medskip
(b) End-to-end 3D reconstruction with estimated conditioning geometry
\smallskip

\resizebox{\columnwidth}{!}{%
\begin{tabular}{ll ccccc}
\toprule
Protocol & Source & Pts $\downarrow$ & Ray$^\circ$ $\downarrow$ & Depth $\downarrow$ & ATE $\downarrow$ & Inl.\ $\uparrow$ \\
\midrule
\multirow{3}{*}{Undist.\ mixed}
 & GT $K$ & \textbf{0.0449} & \textbf{0.161} & \textbf{0.0393} & \textbf{0.0204} & \textbf{81.8} \\
 & AnyCalib $K$ & 0.1161 & 2.543 & 0.1112 & 0.0641 & 50.3 \\
 & VGGT $K$ & 0.0656 & 1.177 & 0.0666 & 0.0366 & 71.0 \\
\midrule
\multirow{2}{*}{Mixed proj.}
 & GT ray map & \textbf{0.0588} & \textbf{0.709} & \textbf{0.0466} & \textbf{0.0238} & \textbf{72.9} \\
 & AnyCalib ray map & 0.0879 & 2.164 & 0.0714 & 0.0514 & 60.2 \\
\bottomrule
\end{tabular}
}
\end{table}

Calibration quality varies across datasets, with the largest focal and principal-point errors on PandaSet.
On undistorted ScanNet++, VGGT estimates yield more accurate 3D reconstruction than single-image AnyCalib estimates, but both trail ground-truth calibration.
On mixed-projection ScanNet++, AnyCalib ray maps achieve 60.2\% point inliers versus 72.9\% with ground-truth ray maps, demonstrating non-pinhole 3D reconstruction with estimated geometry while retaining a measurable calibration gap.
Appendix~\ref{app:estimated_calibration} details the camera models, metric conventions, and per-protocol analysis.

\subsubsection{Runtime and Memory}
\label{subsec:runtime}

\begin{figure}[!t]
  \centering
  \includegraphics[width=\columnwidth]{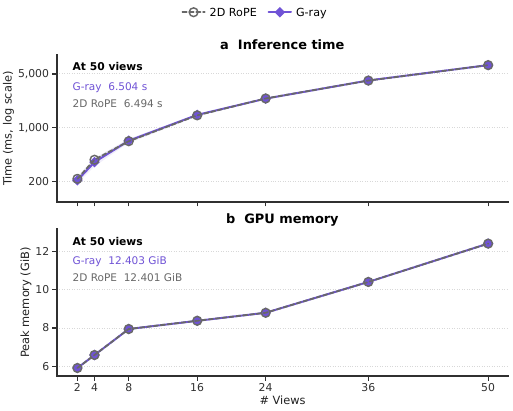}
  \caption{End-to-end inference time and peak GPU memory with 2 to 50 input views. Markers and bands show means and one standard deviation over ten paired trials with alternating execution order on an NVIDIA RTX~4090. Grid-index RoPE and \method{} use the same checkpoint, inference procedure, BF16 precision, and calibrated $518{\times}518$ inputs.}
  \label{fig:runtime_memory}
\end{figure}

We measure the computational cost of coordinate replacement using paired end-to-end inference trials.
Both variants share the same trained checkpoint, prediction heads, and pre- and post-processing.
Only the coordinates supplied to the rotary encoding change between grid indices and calibrated ray angles.
Fig.~\ref{fig:runtime_memory} reports latency and peak GPU memory for 2 to 50 views.
The two modes follow nearly indistinguishable scaling curves.
At 50 views, \method{} and the grid-coordinate baseline respectively use 12.403 and 12.401~GiB on average, while their mean latencies are 6.504 and 6.494~s.
The paired latency difference is 9.6~ms, with a 95\% confidence interval from $-20.0$ to 39.2~ms that includes zero.
These measurements show similar runtime and memory usage, with no detectable latency increase under the evaluated conditions.

\section{Discussion and Limitations}
\label{sec:limitations}

\paragraph{Calibration Dependence}
\method{} requires view-specific image-to-ray geometry, supplied through a calibrated camera model or a unit-ray map.
Tables~\ref{tab:intrinsic_noise} and~\ref{tab:estimated_k} quantify its sensitivity to calibration errors.
Accuracy decreases monotonically across the tested focal perturbations and drops sharply when intrinsics are assigned to the wrong views.
3D reconstruction with either single-image or multi-view estimates remains less accurate than with ground-truth calibration, and estimation quality varies across domains.
These results support using estimated calibration but do not establish calibration-free operation.

\paragraph{Training Scale and Homogeneous-Input Compatibility}
Training data scale and computational resources are important factors in the performance of multi-view vision Transformers.
VGGT-$\Omega$, for example, reports training on 128 H100 GPUs, using approximately 4 million supervised scenes/sequences and 18 million unlabeled videos~\cite{wang2026vggtomega}.
Our 3D reconstruction model is trained on 20 million images from the public-dataset mixture in Appendix~\ref{app:training_reconstruction}, using 8 A100 GPUs.
\method{} addresses cross-projection phase mismatch, a source of inconsistent geometric cues for cross-projection attention, through ray-level angular encoding without additional learned parameters.
With supplied calibration, this model outperforms VGGT-$\Omega$ on all six cross-dataset averaged metrics under camera heterogeneity (Table~\ref{tab:recon_main}) and remains competitive under homogeneous pinhole 3D reconstruction protocols (Table~\ref{tab:pi3_homogeneous}).
These systems differ in architecture, calibration inputs, and training data, so this comparison does not isolate training efficiency.
The results motivate larger-scale training of \method{}, while its performance at a matched data and compute budget remains to be evaluated.

\paragraph{Efficiency Scope}
\method{} introduces no learned parameters.
The paired measurements in Fig.~\ref{fig:runtime_memory} show nearly identical runtime and peak GPU memory scaling for \method{} and grid-index RoPE across 2 to 50 views.
This comparison isolates coordinate replacement for a fixed architecture, input resolution, GPU, and software configuration.
Other architectures, kernels, and hardware require separate benchmarking.

\paragraph{Applicability Boundary}
On Objaverse, the largest gains among the three main regimes occur when novel-viewpoint synthesis is combined with additional FoV change.
For FoV change at a context pose, RayRoPE + \method{} is $0.46$~dB below RayRoPE, while GTA + \method{} improves over GTA.
The target shares one context pose, but the two context views still have different poses.
The benefit therefore depends on the host encoding and target-view regime, and these comparisons do not isolate why RayRoPE's projected pixel coordinates perform better in the observed-pose regime.

The evaluated resolution and projection-model ranges, and the distinction between controlled FoV variation and optical zoom, are discussed further in Appendices~\ref{app:resolution_scope} and~\ref{app:fov_scope}.

\section{Conclusion}
\label{sec:conclusion}

We presented \method{}, a ray-level relative position encoding that lifts the positional coordinate from the image plane to camera-local ray-angle space.
Ray angles provide a shared physical unit across projections, so the resulting relative phase offers projection-invariant positional consistency that image-plane coordinates cannot provide.
The encoding operates by replacing the image-plane coordinate pair used to form rotary phases with ray angles.
The surrounding attention computation and the model's complementary geometric branches are retained.
The method consumes a view-specific image-to-ray map, introduces no learned parameters, and enables a model trained exclusively on pinhole images to generalize to non-pinhole inputs at inference by supplying the corresponding ray map.

Across distributed camera collections, rigid multi-camera rigs, and variable-FoV configurations, \method{} leads all six three-dataset average 3D reconstruction metrics at 50 views among methods with complete results.
NVS experiments show that integrating \method{} into GTA and RayRoPE improves synthesis under controlled FoV variation on RE10K and at novel viewpoints with additional FoV change on Objaverse.
On homogeneous pinhole 3D reconstruction protocols, accuracy is retained under the angular encoding, confirming compatibility when camera heterogeneity is absent.
Further analysis shows that accuracy degrades more gradually than compared baselines under resolution extrapolation beyond the training scale, and that most of it is retained when calibration comes from an off-the-shelf estimator.

The observed calibration sensitivity and the lack of improvement for RayRoPE + \method{} under same-pose FoV changes define current applicability limits and motivate further study of position encoding under camera heterogeneity.

\section*{Acknowledgment}
This work was supported by the National Key R\&D Program of China under Grants 2022ZD0160601 and 2022YFB3903404.

\bibliographystyle{IEEEtran}
\bibliography{references_verified}
\onecolumn
\appendices
\raggedbottom
\section{Method Details}
\label{app:method_details}

\subsection{Image-to-Ray Maps}
\label{app:image_to_ray_maps}

A pinhole camera with zero skew instantiates $\mathcal{U}_c$ through the intrinsic matrix
\begin{equation}
\mathbf{K}_c =
\begin{bmatrix}
f_{x,c} & 0 & c_{x,c} \\
0 & f_{y,c} & c_{y,c} \\
0 & 0 & 1
\end{bmatrix},
\label{eq:intrinsics}
\end{equation}
with $\mathcal{U}_c(\mathbf{p})=\mathbf{K}_c^{-1}\mathbf{p}$.
For the four-coefficient fisheye projection in \eqref{eq:fisheye}, we first normalize the image-plane sample using the intrinsics.
We then numerically invert $\rho(\psi)$ to recover the polar angle $\psi$ from the normalized radius.
Together with the azimuth $\phi$ of the normalized sample, this gives the unit ray $\mathcal{U}_c(\mathbf{p})=(\sin\psi\cos\phi,\,\sin\psi\sin\phi,\,\cos\psi)^{\top}$.
The camera model may also provide a forward projection $\Pi_c$, but \method{} requires only the image-to-ray map.

\subsection{Content-Free Attention Score under Rotary Encoding}
\label{app:pe_only_score}

This subsection derives the content-free (position-only) attention score used in Fig.~\ref{fig:attention} and Sec.~\ref{subsec:attn_diag}.
We first consider a rotary encoding parameterized by coordinates $\boldsymbol{\xi}$, then specialize the result to grid-index RoPE and \method{}.

\subsubsection{Relative Phase of Planar Rotations}
Let $\mathbf{E}(\boldsymbol{\xi})$ denote a block-diagonal bank of planar rotations applied to a feature vector, as in standard RoPE and \eqref{eq:rotation}.
For any query and key features $\mathbf{q},\mathbf{k}\in\mathbb{R}^{d_h}$,
\begin{equation}
\big(\mathbf{E}(\boldsymbol{\xi}_{i})\mathbf{q}\big)^{\!\top}
\big(\mathbf{E}(\boldsymbol{\xi}_{j})\mathbf{k}\big)
=
\mathbf{q}^{\top}
\mathbf{E}(\boldsymbol{\xi}_{i})^{\top}
\mathbf{E}(\boldsymbol{\xi}_{j})
\mathbf{k}.
\label{eq:app_qk}
\end{equation}
Because planar rotations compose by angle addition,
\begin{equation}
\mathbf{E}(\boldsymbol{\xi}_{i})^{\top}
\mathbf{E}(\boldsymbol{\xi}_{j})
=
\mathbf{E}(\boldsymbol{\xi}_{j}-\boldsymbol{\xi}_{i}).
\label{eq:app_compose}
\end{equation}
The positional contribution therefore depends on the relative displacement $\Delta\boldsymbol{\xi}=\boldsymbol{\xi}_{j}-\boldsymbol{\xi}_{i}$ rather than on absolute coordinates.

\subsubsection{Content-Free Probe}
The position-only diagnostic removes image content by setting $\mathbf{q}=\mathbf{k}=\mathbf{1}$ (a constant vector) before applying $\mathbf{E}(\cdot)$.
Consider a two-channel rotation block at frequency $\omega$ with scalar positional coordinate $\xi\in\{\xi^{(1)},\xi^{(2)}\}$.
The two components of $\boldsymbol{\xi}$ are grid-index coordinates for RoPE and $(\theta_x,\theta_y)$ for \method{}.
The corresponding planar rotation acts on a two-dimensional feature pair $(x_1,x_2)$ as
\begin{equation}
R_{\omega\xi}
\begin{bmatrix}x_1\\x_2\end{bmatrix}
=
\begin{bmatrix}
\cos(\omega\xi)\,x_1-\sin(\omega\xi)\,x_2\\
\sin(\omega\xi)\,x_1+\cos(\omega\xi)\,x_2
\end{bmatrix}.
\label{eq:app_planarro}
\end{equation}
Substituting the constant pair $(1,1)$ at positions $i$ and $j$ and taking the inner product yields
\begin{align}
&\left(
R_{\omega\xi_i}
\begin{bmatrix}1\\1\end{bmatrix}
\right)^{\!\top}
\left(
R_{\omega\xi_j}
\begin{bmatrix}1\\1\end{bmatrix}
\right)
\nonumber\\
&\qquad=
\big(\cos\omega\xi_i-\sin\omega\xi_i\big)
\big(\cos\omega\xi_j-\sin\omega\xi_j\big)
\nonumber\\
&\qquad\quad+
\big(\sin\omega\xi_i+\cos\omega\xi_i\big)
\big(\sin\omega\xi_j+\cos\omega\xi_j\big)
\nonumber\\
&\qquad=
2\cos\!\big(\omega(\xi_i-\xi_j)\big).
\label{eq:app_cos}
\end{align}
Summing over frequency channels and over the two axes of $\boldsymbol{\xi}$ therefore gives the content-free score
\begin{equation}
s(i,j)
=
\sum_{a\in\{1,2\}}
\sum_{k}
2\cos\!\big(
\omega_k\,
\big(\xi^{(a)}_{i}-\xi^{(a)}_{j}\big)
\big),
\label{eq:app_score}
\end{equation}
where $\omega_k$ follows the frequency schedule of the evaluated encoding.
Equation~\eqref{eq:app_score} is the raw content-free dot product underlying the PE-only column of Fig.~\ref{fig:attention}.
Attention scaling and normalization are applied afterward, so this expression describes the positional score rather than the final attention weights.

\subsubsection{Specialization to Grid-Index RoPE and \method{}}
Grid-index RoPE uses the token-lattice coordinate pair $\boldsymbol{\xi}=(m,n)$.
When two views share a patch grid, the same lattice offsets produce the same positional scores regardless of focal length or principal point.
The grid-based score therefore does not distinguish their FoVs, even when their visible angular regions differ.
\method{} instead sets $\boldsymbol{\xi}=(\theta_x,\theta_y)$ via Eqs.~\eqref{eq:thetax} and~\eqref{eq:thetay}.
Relative phase then depends on camera-local ray-angle differences.
Identical ray-angle coordinates maximize this position-only score, but its multi-frequency cosine form does not generally decrease monotonically with angular separation.
In the wide/zoom examples of Fig.~\ref{fig:attention}, the resulting prior emphasizes the shared angular region.
The two encodings thus share the cosine mechanism in \eqref{eq:app_score} but differ in the geometric meaning of their positional coordinates.

\subsection{Consistency under Image Resampling}
\label{app:resampling_consistency}

Let $\mathbf{A}$ denote the coordinate transformation that a resize or crop induces on the retained image domain, so that $\mathbf{p}'=\mathbf{A}\mathbf{p}$.
Suppose the updated image-to-ray map satisfies
\begin{equation}
\mathcal{U}'_c(\mathbf{A}\mathbf{p})\propto\mathcal{U}_c(\mathbf{p}),
\label{eq:ray_resampling}
\end{equation}
with a positive proportionality factor.
Then $\mathbf{p}'$ and $\mathbf{p}$ represent the same camera-local ray and receive the same ray angles under \eqref{eq:thetax} and \eqref{eq:thetay}.
For a pinhole camera the same operation updates the intrinsics as $\mathbf{K}'_c=\mathbf{A}\mathbf{K}_c$, so that
\begin{equation}
{\mathbf{K}'_c}^{-1}\mathbf{p}'
=
(\mathbf{A}\mathbf{K}_c)^{-1}\mathbf{A}\mathbf{p}
=
\mathbf{K}_c^{-1}\mathbf{p},
\label{eq:pinhole_resampling}
\end{equation}
which verifies \eqref{eq:ray_resampling} for that case.
This equality concerns exactly corresponding continuous image locations.
Patch centers resampled onto a different discrete grid need not coincide with those locations and generally represent nearby, rather than identical, viewing rays.

\subsection{Positional Encoding Frequency Parameters}
\label{app:freq_params}

\method{} retains each host encoding's allocation of channels to geometric components and uses host-specific frequency settings.
An angular subspace with $n_{\Omega}$ frequency components applies independent two-channel rotations,
\begin{equation}
\mathbf{R}_{\boldsymbol{\Omega}}(\theta)
=
\bigoplus_{k=0}^{n_{\Omega}-1}
\begin{bmatrix}
\cos(\omega_k\theta) & -\sin(\omega_k\theta) \\
\sin(\omega_k\theta) & \cos(\omega_k\theta)
\end{bmatrix}.
\label{eq:rotation_bank}
\end{equation}
For 3D reconstruction and GTA, the frequencies follow \eqref{eq:frequency}, with the number of components determined by the angular subspace.
The frequency scale $\alpha$ and base $\beta$ correspond to \texttt{freq\_scale} and \texttt{freq\_base} in these implementations.
All ray-angle coordinates are expressed in radians.

\paragraph{3D Reconstruction}
The aggregator uses $\alpha{=}1$ and $\beta{=}100$.
Its head channels are divided equally between $\theta_x$ and $\theta_y$, giving $n_{\Omega}{=}d_h/4$ frequencies per angle.
The value and inverse-output transformations in \eqref{eq:output} are enabled.

\paragraph{GTA + \method{}}
The angular subspace uses $\alpha{=}10$ and $\beta{=}100$, while the remaining geometric components retain the host's formulation.
The frequency-scale ablation in Sec.~\ref{subsec:ablation_nvs} varies $\alpha$ with $\beta$ fixed.

\paragraph{RayRoPE + \method{}}
RayRoPE uses a period-based parameterization rather than the scale/base convention above.
For its ray-angle coordinates, the maximum period is $P_{\max}{=}8(\pi/2){=}4\pi$ and the ratio between consecutive frequencies is $b_{\mathrm{R}}{=}3$,
\begin{equation}
\omega^{\mathrm{R}}_k=\frac{2\pi}{P_{\max}}\,b_{\mathrm{R}}^k,
\qquad k=0,\ldots,n_{\Omega}-1.
\label{eq:rayrope_frequency}
\end{equation}
In the implementation, this period is specified by \texttt{max\_period} $=$ \texttt{MAX\_FOV} $\times 8$, with \texttt{MAX\_FOV} $=\pi/2$.
RayRoPE's \texttt{freq\_base} denotes $b_{\mathrm{R}}$, not the decay base $\beta$ in \eqref{eq:frequency}.
The frequency settings for its other geometric components are unchanged.

\subsection{Multiple Positional Samples per Token}
\label{app:multiple_samples}

A host encoding may assign one or more two-dimensional positional samples to each token, denoted by $\boldsymbol{\xi}_{c,i,\ell}$ for sample $\ell$.
Standard 2D RoPE uses a single token-lattice index, whereas a projection-based encoding can provide several pixel locations obtained in a query camera.
\method{} maps each sample independently through \eqref{eq:camera_ray} to \eqref{eq:thetay}, while the host encoding retains its aggregation over $\ell$.
The main text omits the sample index for clarity.

\section{Evaluation Protocols and Metrics}
\label{app:eval_protocols}

\subsection{Objaverse Target-View Definitions}
\label{app:objaverse_protocols}

All six Objaverse protocols evaluate the same 8\,000 test objects under heterogeneous-FoV inputs.
Each object is rendered from eight viewing directions, with an independently sampled FoV and a corresponding camera distance for each native view.
Two native views from different directions are held fixed as context across the six protocols.
For each native view, a focal variant changes its FoV at the same pose, while a dolly variant changes only the camera distance and retains the native FoV.
Here, native FoV denotes the per-view sampled FoV, not a common FoV shared by all views.

The three main regimes in Table~\ref{tab:nvs_splits} differ in the target camera pose and FoV.
\emph{FoV change} uses the two focal variants at the context poses.
\emph{Novel viewpoint} uses the six native views at unseen poses, without further modifying their independently sampled FoVs.
\emph{Novel viewpoint + FoV change} uses the six focal variants at those unseen poses.

The additional protocols include targets with radial camera motion.
\emph{FoV / dolly change} combines two focal and two dolly variants at the context viewing directions, for four targets per object.
\emph{Novel viewpoint + FoV / dolly change} combines six focal and six dolly variants at unseen viewing directions, for twelve targets.
\emph{Original mixed protocol} combines the four FoV/dolly targets with the six native novel-viewpoint targets, for ten targets.
It excludes the focal and dolly variants at unseen viewing directions and is not an average over all six protocols.
The FoV/dolly protocols mix changes in FoV and camera distance, rather than isolating either factor.

\subsection{Inlier and Pose Metric Definitions}
\label{app:metric_definitions}

For a ground-truth point $\mathbf q$ and prediction $\widehat{\mathbf q}$, the point-inlier test is
\begin{equation}
\max\!\left(\frac{\|\widehat{\mathbf q}\|_2}{\|\mathbf q\|_2},
             \frac{\|\mathbf q\|_2}{\|\widehat{\mathbf q}\|_2}\right)<1.03.
\label{eq:inlier_ratio}
\end{equation}
The evaluator first expresses points in the first-view camera frame and independently normalizes prediction and ground truth by their mean valid point distance over each input set.
Depth inliers apply the same ratio test to scalar depths, using $z$-depth for PandaSet and ETH3D and ray depth for mixed-projection ScanNet++.
Both rates are averaged over valid pixels, then views and evaluation sets, and reported in percent.
Thus, $1.03$ is a dimensionless threshold; the point test compares point norms rather than Euclidean point error.
The main reconstruction pose ATE is the mean Euclidean translation error after rigid trajectory alignment in these normalized coordinates.

\subsection{NVS Metric Aggregation}
\label{app:nvs_aggregation}

All NVS metrics use RGB intensities in $[0,1]$ without foreground masking.
For scene or object $s$ with $T_s$ target images, let $e_{s,t}$ be the MSE over the pixels and color channels of target $t$.
The reported PSNR is
\begin{equation}
\operatorname{PSNR}=\frac{1}{S}\sum_{s=1}^{S}
\left[-10\log_{10}\left(\frac{1}{T_s}\sum_{t=1}^{T_s}e_{s,t}\right)\right].
\label{eq:nvs_psnr_aggregation}
\end{equation}
SSIM and AlexNet LPIPS are averaged over target images within each scene or object, followed by an equal-weight scene/object average.
Evaluation uses three targets for each of the 6,202 RE10K scenes and protocol-specific target sets for the 8,000 Objaverse objects.
The original mixed Objaverse protocol pools four FoV/dolly and six native novel-viewpoint targets per object, so its PSNR is computed from their pooled MSE before the logarithm.
Consequently, the mixed PSNR is not a weighted mean of the component PSNR columns.

\section{Training Details}
\label{app:training_details}

\subsection{3D Reconstruction}
\label{app:training_reconstruction}

The 3D reconstruction model uses a VGGT-based architecture~\cite{wang2025vggt} with a DINOv2 ViT-L/14 backbone (embedding dimension 1024, 24 layers, 16 heads, and patch size 14) and 24 pairs of alternating frame and global attention blocks.
A DPT head predicts camera-space pointmaps, and a linear patch head predicts world-space pointmaps.
A separate pose-regression head processes the final patch features with residual convolutions and global average pooling, followed by an MLP with translation and quaternion output branches.

The training mixture comprises ARKitScenes~\cite{baruch2021arkitscenes}, BlendedMVS~\cite{yao2020blendedmvs}, Habitat-Matterport 3D (HM3D)~\cite{ramakrishnan2021hm3d}, Hypersim~\cite{roberts2021hypersim}, MatrixCity~\cite{li2023matrixcity}, MegaDepth~\cite{li2018megadepth}, Mid-Air~\cite{fonder2019midair}, Matterport3D~\cite{chang2017matterport3d}, OmniObject3D~\cite{wu2023omniobject3d}, OmniWorld-Game~\cite{zhou2025omniworld}, Aria Synthetic Environments (ASE)~\cite{avetisyan2024scenescript}, ScanNet++~\cite{yeshwanth2023scannetpp}, Taskonomy~\cite{zamir2018taskonomy}, UnrealStereo4K~\cite{tosi2021smdnets}, WildRGB-D~\cite{xia2024wildrgbd}, Dynamic Replica~\cite{karaev2023dynamicstereo}, MVS-Synth~\cite{huang2018deepmvs}, TartanAir~\cite{wang2020tartanair}, Virtual KITTI 2~\cite{cabon2020virtualkitti2}, and the Waymo Open Dataset~\cite{sun2020waymo}.

We initialize the image encoder with pretrained DINOv2 weights and train the 3D reconstruction model in two stages.
Both stages use 8 NVIDIA A100 GPUs with distributed data parallelism.
\textbf{Stage~1} trains for 120 epochs at a 224-pixel long side using AdamW with a peak learning rate of $5{\times}10^{-5}$ and weight decay of 0.05.
The schedule uses 5\% linear warmup from $10^{-8}$ followed by cosine decay to $10^{-8}$.
Stage~1 uses BF16 mixed precision, gradient clipping at max-norm 1.0, and up to 240 images per GPU.
Each batch samples 2 to 24 views per sequence and a shared aspect ratio between 0.33 and 1.0, with the short side aligned to the 14-pixel patch size.

ScanNet++ also appears in the training corpora reported for $\pi^3$ and MapAnything~\cite{wang2025pi3,keetha2025mapanything}.
Here, pinhole-to-non-pinhole transfer refers to the change in imaging model within the evaluation protocol, rather than an unseen dataset domain.
The 50 ScanNet++ evaluation scenes follow its validation split, which is disjoint from the official training scene list.

\textbf{Stage~2} fine-tunes the Stage-1 checkpoint for 60 epochs at a 518-pixel long side.
It uses an initial learning rate of $10^{-7}$, cosine decay to $10^{-10}$ without warmup, and up to 48 images per GPU.
Both stages are configured for 800 gradient-update steps per epoch without gradient accumulation, corresponding to approximately 96\,k and 48\,k updates, respectively.

The training objective is a weighted sum of
(i)~a camera loss ($\lambda{=}5.0$) computing L1 error on the final predicted absolute translation and quaternion rotation,
(ii)~a camera-space point loss ($\lambda{=}1.0$) with confidence-weighted L2 regression and a normal-based gradient loss after outlier filtering at the 98th percentile, and
(iii)~a world-space point loss ($\lambda{=}1.0$) using the same formulation.
Ground truth is scale-normalized by the mean world-point distance.

\subsection{Novel View Synthesis}
\label{app:training_nvs}

All NVS variants share the LVSM decoder-only backbone~\cite{jin2024lvsm}, with 6 transformer layers, model dimension 1152, 8 heads, feed-forward dimension 1024, and patch size 8.
Training uses 2 context views and 1 target view at $256{\times}256$ resolution.
Models are trained separately on RE10K and Objaverse.
The Objaverse training set contains approximately 72\,k objects, each rendered from 8 viewing directions with additional FoV and camera-distance variants.

Each positional encoding variant is trained independently on a single NVIDIA RTX~4090 GPU with batch size 4 and FP16 mixed precision.
We use AdamW with a learning rate of $4{\times}10^{-4}$, $\beta_1{=}0.9$, $\beta_2{=}0.95$, and weight decay 0.5.
Training runs for 80\,k steps, with 2\,500 steps of linear warmup followed by cosine decay.

The loss is $\mathcal{L}=\mathcal{L}_\text{MSE}+0.5\,\mathcal{L}_\text{LPIPS}$, where $\mathcal{L}_\text{MSE}$ is the pixel-wise mean squared error between the predicted and ground-truth target view, and $\mathcal{L}_\text{LPIPS}$ is the AlexNet-based perceptual loss~\cite{zhang2018lpips}.
No explicit depth, pose, or adversarial losses are used.
In RayRoPE, per-layer depth and uncertainty predictions are used within the positional encoding and receive no separate supervision.

\section{Per-Dataset 3D Reconstruction Results}
\label{app:recon_per_dataset}

Table~\ref{tab:recon_per_dataset} reports the per-dataset 50-view results used to compute the three-dataset averages in Table~\ref{tab:recon_main}.
Method publication details are given in Table~\ref{tab:recon_main}.

\begin{table}[H]
\centering
\footnotesize
\setlength{\tabcolsep}{2.0pt}
\caption{Per-dataset 3D reconstruction results at 50 input views. Mixed-projection ScanNet++ combines non-pinhole DSLR and pinhole iPhone views and uses ray-depth metrics. The w/C group receives external calibration through each model's supported interface; WorldMirror uses pinhole intrinsics, while the other three methods receive calibrated ray maps. The w/o C group receives no external calibration and includes image-only methods and \method{} with AnyCalib-estimated ray maps. Bold and underlined values indicate the best and second-best results for each metric within each dataset.}
\label{tab:recon_per_dataset}
\begin{tabular}{lll cccccc}
\toprule
Dataset & Setting & Method & Points rel $\downarrow$ & Rays err$^\circ$ $\downarrow$ & Depth rel $\downarrow$ & Pts Inl.\ $\uparrow$ & Pose ATE $\downarrow$ & Dp.\ Inl.\ $\uparrow$ \\
\midrule
\multirow{13}{*}{\shortstack[l]{ScanNet++\\{\scriptsize (dist. fish.+pin.)}}}
& \multirow{4}{*}{w/C} & \method{} & \textbf{0.0583} & \textbf{0.72} & \textbf{0.0449} & \textbf{75.1} & \textbf{0.0246} & \textbf{66.0} \\
& & $\pi^3$-X & 0.2080 & 4.29 & 0.1052 & 25.8 & 0.1005 & 21.8 \\
& & WorldMirror & 0.2347 & 4.16 & 0.1010 & 27.4 & 0.0969 & 22.9 \\
& & MapAnything & 0.2246 & 3.39 & 0.1729 & 26.2 & 0.1312 & 9.6 \\
\cmidrule(l){2-9}
& \multirow{9}{*}{w/o C} & \method{} + AnyCalib & \underline{0.0872} & \underline{2.18} & \underline{0.0707} & \underline{61.4} & \underline{0.0518} & \underline{36.7} \\
& & DA3 & 0.4646 & 9.94 & 0.2501 & 14.2 & 0.2152 & 5.1 \\
& & DA3-Nested & 0.4549 & 9.58 & 0.2367 & 15.5 & 0.2012 & 5.9 \\
& & MapAnything & 0.2806 & 5.28 & 0.1906 & 24.1 & 0.1536 & 7.8 \\
& & $\pi^3$ & 0.2846 & 6.95 & 0.1724 & 22.4 & 0.1457 & 7.0 \\
& & VGGT & 0.4260 & 7.96 & 0.1942 & 14.2 & 0.2068 & 9.6 \\
& & VGGT-$\Omega$ & 0.4133 & 7.83 & 0.1736 & 16.3 & 0.1556 & 9.9 \\
& & OmniVGGT & 0.2035 & 3.98 & 0.1225 & 34.1 & 0.1130 & 12.5 \\
& & WorldMirror & 0.3306 & 5.78 & 0.1405 & 20.9 & 0.1349 & 16.5 \\
\midrule
\multirow{13}{*}{PandaSet}
& \multirow{4}{*}{w/C} & \method{} & \underline{0.1099} & \textbf{0.27} & \underline{0.1378} & 34.7 & \underline{0.0117} & \underline{27.5} \\
& & $\pi^3$-X & \textbf{0.0959} & \underline{2.33} & \textbf{0.1115} & \textbf{45.7} & \textbf{0.0106} & \textbf{28.1} \\
& & WorldMirror & 0.1264 & 2.39 & 0.1508 & 21.7 & 0.0127 & 19.2 \\
& & MapAnything & 0.1233 & 2.58 & 0.1406 & \underline{35.8} & 0.0171 & 26.5 \\
\cmidrule(l){2-9}
& \multirow{9}{*}{w/o C} & \method{} + AnyCalib & 0.2191 & 3.71 & 0.2607 & 22.0 & 0.0820 & 10.8 \\
& & DA3 & 0.4725 & 8.47 & 0.5302 & 12.2 & 0.2980 & 7.3 \\
& & DA3-Nested & 0.4172 & 6.45 & 0.4039 & 13.9 & 0.1591 & 5.0 \\
& & MapAnything & 0.3245 & 4.86 & 0.3325 & 22.0 & 0.1349 & 8.4 \\
& & $\pi^3$ & 0.5000 & 4.15 & 0.3355 & 18.9 & 0.2549 & 9.2 \\
& & VGGT & 0.4241 & 4.07 & 0.3604 & 11.1 & 0.2161 & 9.1 \\
& & VGGT-$\Omega$ & 0.1676 & 4.04 & 0.2098 & 30.6 & 0.0680 & 15.6 \\
& & OmniVGGT & 0.4339 & 7.18 & 0.7329 & 6.3 & 0.2925 & 4.0 \\
& & WorldMirror & 0.1854 & 3.42 & 0.2717 & 20.0 & 0.0767 & 13.1 \\
\midrule
\multirow{13}{*}{ETH3D}
& \multirow{4}{*}{w/C} & \method{} & 0.0549 & \textbf{0.15} & 0.0405 & 64.6 & 0.0300 & 62.2 \\
& & $\pi^3$-X & \underline{0.0352} & 0.33 & \underline{0.0228} & \underline{74.7} & \underline{0.0163} & \underline{76.9} \\
& & WorldMirror & 0.0401 & \underline{0.24} & 0.0290 & 71.8 & 0.0227 & 72.5 \\
& & MapAnything & 0.0636 & 0.35 & 0.0461 & 50.4 & 0.0365 & 49.7 \\
\cmidrule(l){2-9}
& \multirow{9}{*}{w/o C} & \method{} + AnyCalib & 0.0977 & 1.56 & 0.0908 & 46.8 & 0.0767 & 25.6 \\
& & DA3 & 0.0952 & 1.82 & 0.0679 & 49.5 & 0.0524 & 38.4 \\
& & DA3-Nested & 0.1113 & 2.11 & 0.0756 & 47.7 & 0.0730 & 36.4 \\
& & MapAnything & 0.1162 & 1.57 & 0.0879 & 34.9 & 0.0738 & 25.8 \\
& & $\pi^3$ & 0.0638 & 1.79 & 0.0738 & 63.0 & 0.0603 & 47.8 \\
& & VGGT & 0.1575 & 0.90 & 0.0699 & 43.1 & 0.1155 & 39.6 \\
& & VGGT-$\Omega$ & \textbf{0.0225} & 0.49 & \textbf{0.0152} & \textbf{91.6} & \textbf{0.0105} & \textbf{88.7} \\
& & OmniVGGT & 0.1750 & 1.74 & 0.1038 & 36.9 & 0.1388 & 25.1 \\
& & WorldMirror & 0.0672 & 0.82 & 0.0541 & 55.0 & 0.0400 & 45.7 \\
\bottomrule
\end{tabular}
\end{table}

\paragraph{Three-Dataset Average}
With supplied calibration, \method{} leads all six cross-dataset metrics at 50 views, including a pointmap relative error of 0.0744, a ray-direction error of $0.38^\circ$, and 58.1\% point inliers.
$\pi^3$-X ranks second on five metrics, while MapAnything ranks second on ray-direction error.
Without externally supplied calibration, \method{} + AnyCalib achieves the lowest average pointmap error (0.1347), ray-direction error ($2.49^\circ$), and pose ATE (0.0702).
VGGT-$\Omega$ remains strongest in this group on depth error and point and depth inliers.
The largest gains over image-only methods occur on mixed-projection ScanNet++, while the other datasets reveal complementary strengths of the baselines.

\paragraph{ScanNet++ (Mixed Projections, Ray Depth)}
This protocol tests generalization from pinhole training images to mixed pinhole and non-pinhole inputs through the calibrated ray-map interface.
With ground-truth ray maps, \method{} achieves the best result on every reported metric, including 0.0583 pointmap relative error, 0.0449 ray-depth relative error, and 75.1\% point inliers.
Among the other calibrated methods, $\pi^3$-X and WorldMirror improve over MapAnything on several depth and pose metrics but remain less accurate than \method{} on pointmaps and ray depth.
Replacing ground-truth calibration with AnyCalib estimates yields 0.0872 pointmap error and 61.4\% point inliers.
This estimated-calibration variant ranks second across all six ScanNet++ metrics, behind \method{} with ground-truth calibration and ahead of every image-only baseline.

\paragraph{PandaSet (Rigid Multi-Camera Rig)}
$\pi^3$-X leads most metrics on PandaSet, while \method{} achieves the lowest ray-direction error ($0.27^\circ$) and ranks second on pointmap error, depth error, pose ATE, and depth inliers.
WorldMirror with supplied calibration improves over its image-only counterpart, but trails \method{} and $\pi^3$-X on the primary error measures.
Unlike mixed-projection ScanNet++, PandaSet uses a rigid rig with a fixed imaging geometry for each camera.
The results show that the advantage of \method{} depends on the evaluation setting and does not extend to every metric of every heterogeneous rig.

\paragraph{ETH3D (Controlled FoV Variation)}
VGGT-$\Omega$ leads most geometry metrics on ETH3D.
\method{} achieves the lowest ray-direction error ($0.15^\circ$).
Among methods with supplied calibration, $\pi^3$-X and WorldMirror are strong, with MapAnything trailing on all six ETH3D columns.
Here, center cropping induces FoV differences within a shared pinhole projection model.
The gains are concentrated in ray-direction accuracy, rather than a uniform improvement across all 3D reconstruction metrics.

\section{Homogeneous Pinhole 3D Reconstruction Protocols}
\label{app:pi3_protocol}

This appendix reports complete relative-pose, video-depth, and multi-view 3D reconstruction results under the homogeneous pinhole 3D reconstruction protocols~\cite{wang2025pi3} summarized in Sec.~\ref{subsec:pi3_protocol}.
The undistorted ETH3D split here differs from the ETH3D setting with controlled FoV variation in Table~\ref{tab:recon_main}.
Accuracy and completeness are measured in millimetres on DTU and in metres on the other 3D reconstruction datasets.
Normal consistency is dimensionless.
KITTI video-depth evaluation uses 64-frame windows for MapAnything and complete sequences for the other methods.
Method publication details are given in Table~\ref{tab:pi3_homogeneous}.

\begin{table}[H]
\centering
\small
\setlength{\tabcolsep}{4.0pt}
\caption{Relative-pose accuracy on TUM and Sintel under homogeneous pinhole 3D reconstruction protocols. ATE and translational RPE use each dataset's distance units, while rotational RPE is in degrees. Lower is better for all metrics. Bold and underlined values indicate the best and second-best results in each column.}
\label{tab:pi3_relpose}
\begin{tabular}{l ccc ccc c}
\toprule
Method & \multicolumn{3}{c}{TUM} & \multicolumn{3}{c}{Sintel} & Avg ATE $\downarrow$ \\
\cmidrule(lr){2-4}\cmidrule(lr){5-7}
 & ATE $\downarrow$ & RPE$_t$ $\downarrow$ & RPE$_r$ $\downarrow$ & ATE $\downarrow$ & RPE$_t$ $\downarrow$ & RPE$_r$ $\downarrow$ & \\
\midrule
\method{} & 0.0122 & 0.0113 & 0.364 & \underline{0.110} & 0.0598 & \underline{0.406} & \underline{0.061} \\
MapAnything & 0.0179 & 0.0187 & 0.359 & 0.219 & 0.0747 & 0.656 & 0.119 \\
VGGT & \underline{0.0090} & \underline{0.0095} & \underline{0.314} & 0.171 & \underline{0.0582} & 0.449 & 0.090 \\
VGGT-$\Omega$ & \textbf{0.0044} & \textbf{0.0046} & \textbf{0.274} & \textbf{0.040} & \textbf{0.0255} & \textbf{0.232} & \textbf{0.022} \\
\bottomrule
\end{tabular}
\end{table}

\begin{table}[H]
\centering
\small
\setlength{\tabcolsep}{3.5pt}
\caption{Video-depth accuracy on Sintel and KITTI under homogeneous pinhole 3D reconstruction protocols. Lower is better for AbsRel, SqRel, and RMSE, while higher is better for $\delta{<}1.25$. Bold and underlined values indicate the best and second-best results in each column.}
\label{tab:pi3_videodepth}
\begin{tabular}{l cccc cccc}
\toprule
Method & \multicolumn{4}{c}{Sintel} & \multicolumn{4}{c}{KITTI} \\
\cmidrule(lr){2-5}\cmidrule(lr){6-9}
 & AbsRel $\downarrow$ & SqRel $\downarrow$ & RMSE $\downarrow$ & $\delta{<}1.25$ $\uparrow$ & AbsRel $\downarrow$ & SqRel $\downarrow$ & RMSE $\downarrow$ & $\delta{<}1.25$ $\uparrow$ \\
\midrule
\method{} & 0.357 & \underline{3.74} & 6.20 & 0.560 & 0.092 & 1.66 & 5.52 & 0.943 \\
MapAnything & 0.401 & 4.01 & \underline{5.55} & \underline{0.620} & 0.083 & 0.774 & \underline{3.97} & \underline{0.946} \\
VGGT & \underline{0.324} & \textbf{3.02} & 6.20 & 0.612 & \underline{0.082} & \underline{0.710} & 4.46 & 0.933 \\
VGGT-$\Omega$ & \textbf{0.251} & 4.59 & \textbf{4.90} & \textbf{0.834} & \textbf{0.072} & \textbf{0.638} & \textbf{3.88} & \textbf{0.952} \\
\bottomrule
\end{tabular}
\end{table}

\begin{table}[H]
\centering
\small
\setlength{\tabcolsep}{2.4pt}
\caption{Indoor multi-view 3D reconstruction under homogeneous pinhole 3D reconstruction protocols. Accuracy (Acc) and completeness (Comp) are measured in metres, with lower values better. Higher normal consistency (NC) is better. Bold and underlined values indicate the best and second-best results for each metric and dataset.}
\label{tab:pi3_mv_indoor}
\begin{tabular}{l ccc ccc ccc ccc}
\toprule
Method & \multicolumn{3}{c}{NRGBD-sparse} & \multicolumn{3}{c}{NRGBD-dense} & \multicolumn{3}{c}{7-Scenes-sparse} & \multicolumn{3}{c}{7-Scenes-dense} \\
\cmidrule(lr){2-4}\cmidrule(lr){5-7}\cmidrule(lr){8-10}\cmidrule(lr){11-13}
 & Acc & Comp & NC & Acc & Comp & NC & Acc & Comp & NC & Acc & Comp & NC \\
\midrule
\method{} & \underline{0.037} & \underline{0.041} & \underline{0.899} & \underline{0.023} & 0.020 & \underline{0.870} & \underline{0.038} & \underline{0.037} & \textbf{0.772} & \underline{0.019} & 0.023 & \underline{0.685} \\
MapAnything & 0.088 & 0.100 & 0.852 & 0.036 & 0.024 & 0.850 & 0.059 & 0.120 & 0.722 & 0.023 & \underline{0.022} & 0.672 \\
VGGT & 0.058 & 0.073 & 0.871 & 0.029 & \underline{0.019} & 0.857 & 0.055 & 0.066 & 0.724 & 0.025 & 0.023 & 0.670 \\
VGGT-$\Omega$ & \textbf{0.022} & \textbf{0.018} & \textbf{0.918} & \textbf{0.014} & \textbf{0.008} & \textbf{0.882} & \textbf{0.029} & \textbf{0.032} & \underline{0.766} & \textbf{0.018} & \textbf{0.018} & \textbf{0.688} \\
\bottomrule
\end{tabular}
\end{table}

\begin{table}[H]
\centering
\small
\setlength{\tabcolsep}{5.0pt}
\caption{3D reconstruction on DTU and undistorted ETH3D under homogeneous pinhole 3D reconstruction protocols. Accuracy (Acc) and completeness (Comp) are measured in millimetres on DTU and metres on ETH3D, with lower values better. Higher normal consistency (NC) is better. Bold and underlined values indicate the best and second-best results in each column. This ETH3D split differs from the controlled FoV variation setting in Table~\ref{tab:recon_main}.}
\label{tab:pi3_mv_dtu_eth3d}
\begin{tabular}{l ccc ccc}
\toprule
Method & \multicolumn{3}{c}{DTU (mm)} & \multicolumn{3}{c}{ETH3D (m)} \\
\cmidrule(lr){2-4}\cmidrule(lr){5-7}
 & Acc $\downarrow$ & Comp $\downarrow$ & NC $\uparrow$ & Acc $\downarrow$ & Comp $\downarrow$ & NC $\uparrow$ \\
\midrule
\method{} & 3.16 & 2.12 & \textbf{0.708} & 0.303 & 0.329 & 0.831 \\
MapAnything & \underline{2.22} & \underline{1.91} & \underline{0.687} & 0.490 & 0.578 & 0.804 \\
VGGT & 2.61 & 2.14 & \underline{0.687} & \underline{0.277} & \underline{0.283} & \underline{0.849} \\
VGGT-$\Omega$ & \textbf{1.22} & \textbf{1.89} & 0.680 & \textbf{0.074} & \textbf{0.062} & \textbf{0.923} \\
\bottomrule
\end{tabular}
\end{table}

\section{Additional NVS Evaluation}
\label{app:additional_nvs}

\subsection{Additional Objaverse Results}
\label{app:objaverse_results}

Baseline references are given in Tables~\ref{tab:nvs_re10k} and~\ref{tab:nvs_splits}.

\begin{table}[H]
\centering
\caption{Additional Objaverse NVS protocols with the same objects, context views, and checkpoints as Table~\ref{tab:nvs_splits}. The mixed protocol contains four FoV/dolly and six native novel-viewpoint targets per object. PSNR pools target MSE within each object before the logarithm; it is not a weighted mean of the other PSNR columns. Bold indicates the best unrounded value.}
\label{tab:nvs_objaverse_additional}
\footnotesize
\setlength{\tabcolsep}{3pt}
\begin{tabular*}{\textwidth}{@{\extracolsep{\fill}}l rrr rrr rrr @{}}
\toprule
Method & \multicolumn{3}{c}{FoV / dolly change} & \multicolumn{3}{c}{Novel viewpoint + FoV / dolly} & \multicolumn{3}{c}{Original mixed protocol} \\
\cmidrule(lr){2-4}\cmidrule(lr){5-7}\cmidrule(lr){8-10}
 & PSNR$\uparrow$ & SSIM$\uparrow$ & LPIPS$\downarrow$ & PSNR$\uparrow$ & SSIM$\uparrow$ & LPIPS$\downarrow$ & PSNR$\uparrow$ & SSIM$\uparrow$ & LPIPS$\downarrow$ \\
\midrule
Pl\"ucker & 26.43 & 0.915 & 0.098 & 13.14 & 0.796 & 0.310 & 20.52 & 0.884 & 0.157 \\
PRoPE & 26.41 & 0.914 & 0.092 & 12.80 & 0.794 & 0.309 & 21.21 & 0.887 & 0.143 \\
\midrule
GTA & 26.11 & 0.911 & 0.097 & 12.97 & 0.796 & 0.308 & 20.68 & 0.883 & 0.154 \\
GTA + \textbf{\method{}} & 26.62 & 0.917 & 0.083 & 13.72 & 0.803 & 0.288 & 21.13 & 0.887 & 0.142 \\
\midrule
RayRoPE & \textbf{28.19} & \textbf{0.935} & \textbf{0.057} & 12.85 & 0.796 & 0.309 & 21.32 & 0.895 & 0.131 \\
RayRoPE + \textbf{\method{}} & 28.01 & 0.931 & 0.063 & \textbf{14.23} & \textbf{0.820} & \textbf{0.254} & \textbf{21.68} & \textbf{0.896} & \textbf{0.123} \\
\bottomrule
\end{tabular*}
\end{table}

Table~\ref{tab:nvs_objaverse_additional} reports results under these additional protocols.
GTA + \method{} improves over GTA in all three protocols.
RayRoPE + \method{} improves by 1.38~dB in Novel viewpoint + FoV / dolly change and by 0.36~dB in Original mixed protocol, but is 0.18~dB lower in FoV / dolly change.
Together with the main results, these comparisons show that gains depend on the target-view setting.

\clearpage
\subsection{Additional NVS Visualizations}
\label{app:nvs_visualizations}

\begin{figure}[H]
  \centering
  \includegraphics[width=\textwidth]{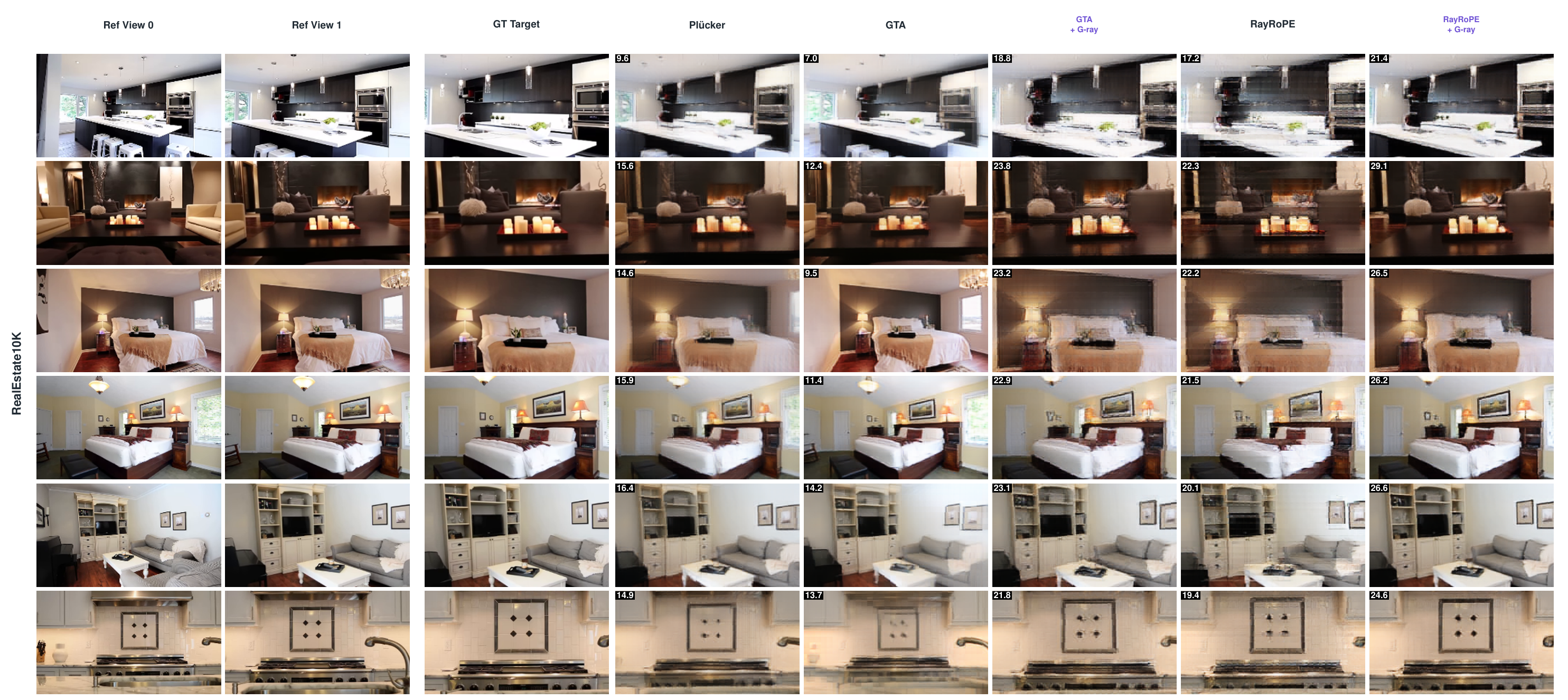}
  \caption{Additional RE10K NVS examples under controlled FoV variation with $z_{\max}{=}2.0$. Rows show different scenes. From left to right, columns contain two context views, the ground-truth target, and predictions from Pl\"ucker, GTA, GTA + \method{}, RayRoPE, and RayRoPE + \method{}. Numbers on predictions report target-image PSNR (dB $\uparrow$). The paired host-encoding columns allow comparison of structural alignment, blur, and ghosting.}
  \label{fig:supp_nvs_re10k}
\end{figure}

\begin{figure}[H]
  \centering
  \includegraphics[width=\textwidth]{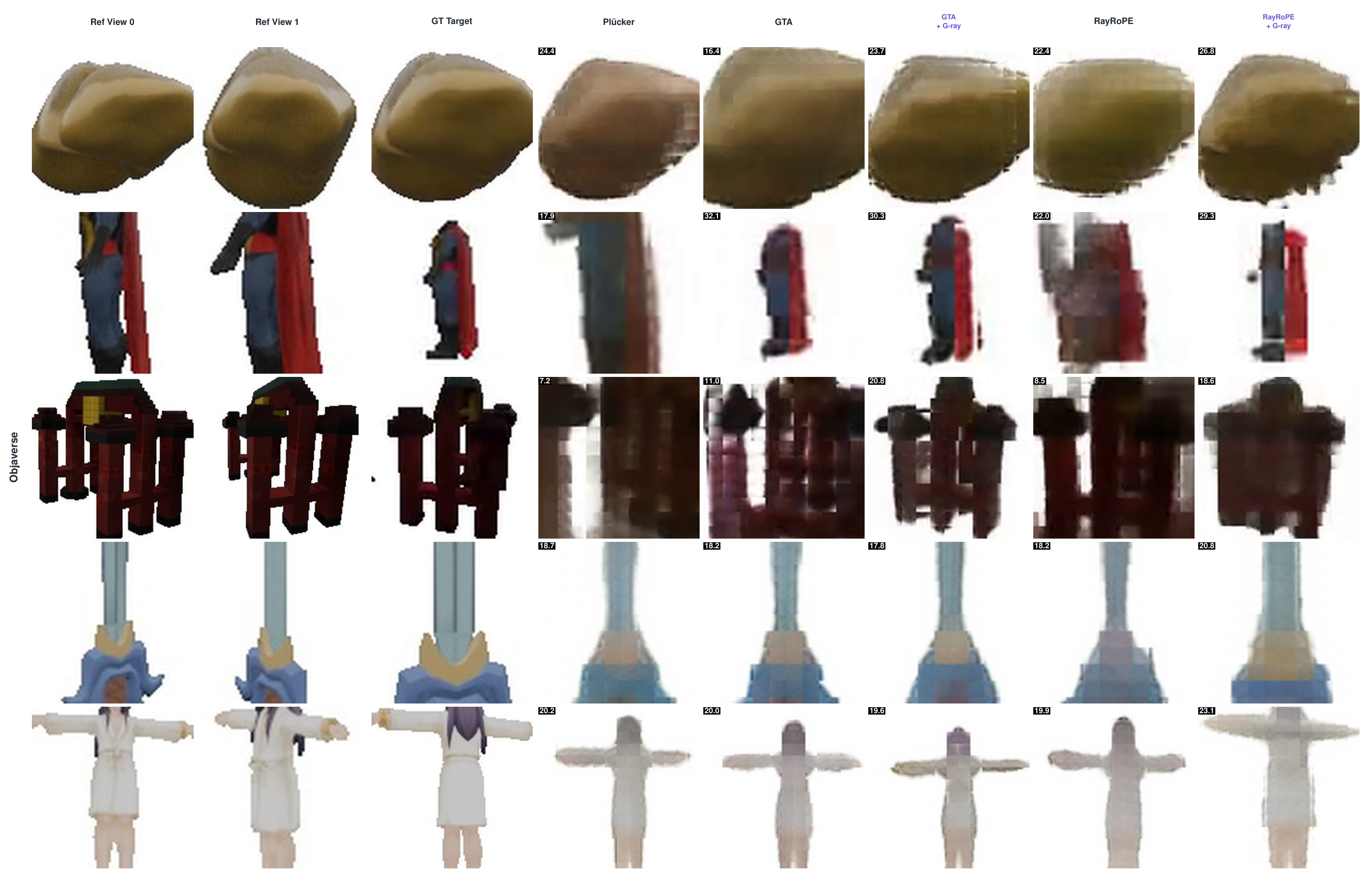}
  \caption{Additional Objaverse NVS examples under FoV heterogeneity. Rows show different objects. From left to right, columns contain two context views, the ground-truth target, and predictions from Pl\"ucker, GTA, GTA + \method{}, RayRoPE, and RayRoPE + \method{}. Numbers on predictions report target-image PSNR (dB $\uparrow$). These examples illustrate changes in object shape and appearance; the gains from ray-angle coordinates vary across targets.}
  \label{fig:supp_nvs_objaverse}
\end{figure}

Figures~\ref{fig:supp_nvs_re10k} and~\ref{fig:supp_nvs_objaverse} provide additional NVS comparisons on RE10K~\cite{zhou2018realestate10k} and Objaverse~\cite{deitke2023objaverse}, respectively.
Each row contains two context views, a ground-truth target, and predictions from the Pl\"ucker baseline~\cite{jin2024lvsm}, GTA~\cite{miyato2023gta}, GTA + \method{}, RayRoPE~\cite{wu2026rayrope}, and RayRoPE + \method{}.
The paired columns show how ray-angle coordinates affect scene structure and texture in individual examples.
The overlaid PSNR values measure individual target images and complement the aggregate results in Tables~\ref{tab:nvs_re10k} and~\ref{tab:nvs_splits}.

\clearpage
\clearpage

\subsection{Complete NVS Frequency-Scale Results}
\label{app:nvs_frequency}

Table~\ref{tab:nvs_frequency_full} complements Table~\ref{tab:ablation_fs} with SSIM and LPIPS for the same frequency-scale sweep.
The selected $\alpha{=}10$ gives the lowest LPIPS in all four protocols; PSNR and SSIM favor different scales in some settings.

\begin{table}[H]
\centering
\caption{Complete frequency-scale sweep for GTA + \method{} on RE10K at $256{\times}144$, with $\beta{=}100$. Each setting uses 6,202 scenes and one trained seed. Bold marks the best unrounded value. All metrics use the evaluation protocols in Table~\ref{tab:nvs_re10k}.}
\label{tab:nvs_frequency_full}
\footnotesize
\setlength{\tabcolsep}{3pt}
\begin{tabular*}{\textwidth}{@{\extracolsep{\fill}}r rrr rrr rrr rrr@{}}
\toprule
$\alpha$ & \multicolumn{3}{c}{$z{=}1$} & \multicolumn{3}{c}{$z_{\max}{=}1.5$} & \multicolumn{3}{c}{$z_{\max}{=}2.0$} & \multicolumn{3}{c}{$z_{\max}{=}2.5$} \\
\cmidrule(lr){2-4}\cmidrule(lr){5-7}\cmidrule(lr){8-10}\cmidrule(lr){11-13}
 & PSNR$\uparrow$ & SSIM$\uparrow$ & LPIPS$\downarrow$ & PSNR$\uparrow$ & SSIM$\uparrow$ & LPIPS$\downarrow$ & PSNR$\uparrow$ & SSIM$\uparrow$ & LPIPS$\downarrow$ & PSNR$\uparrow$ & SSIM$\uparrow$ & LPIPS$\downarrow$ \\
\midrule
1 & 20.66 & 0.642 & 0.201 & 20.40 & 0.607 & 0.213 & 19.98 & 0.581 & 0.240 & 19.65 & 0.564 & 0.266 \\
2 & \textbf{20.97} & \textbf{0.656} & 0.191 & \textbf{20.74} & 0.625 & 0.199 & 20.33 & 0.599 & 0.223 & 20.01 & 0.583 & 0.248 \\
4 & 20.77 & 0.647 & 0.200 & 20.63 & 0.618 & 0.211 & 20.26 & 0.594 & 0.236 & 19.98 & 0.579 & 0.262 \\
6 & 20.76 & 0.646 & 0.204 & 20.53 & 0.615 & 0.218 & 20.08 & 0.589 & 0.246 & 19.72 & 0.572 & 0.274 \\
8 & 20.54 & 0.643 & 0.194 & 20.58 & 0.620 & 0.196 & 20.30 & 0.598 & 0.218 & 20.04 & 0.581 & 0.244 \\
10 & 20.62 & 0.651 & \textbf{0.178} & 20.72 & 0.632 & \textbf{0.181} & \textbf{20.41} & 0.608 & \textbf{0.204} & \textbf{20.12} & 0.588 & \textbf{0.230} \\
15 & 20.31 & 0.635 & 0.193 & 20.50 & 0.621 & 0.194 & 20.23 & 0.599 & 0.216 & 19.96 & 0.582 & 0.241 \\
20 & 20.39 & 0.641 & 0.187 & 20.51 & 0.622 & 0.195 & 20.19 & 0.598 & 0.220 & 19.90 & 0.581 & 0.245 \\
40 & 20.25 & 0.641 & 0.205 & 20.60 & \textbf{0.633} & 0.204 & 20.31 & \textbf{0.610} & 0.224 & 20.01 & \textbf{0.592} & 0.248 \\
\bottomrule
\end{tabular*}
\end{table}

\section{Calibration and Evaluation Scope}
\label{app:evaluation_scope}

\subsection{Off-the-Shelf Intrinsic Estimates}
\label{app:estimated_calibration}

We separately evaluate AnyCalib, a single-image estimator that regresses per-pixel viewing rays and recovers model-specific calibration~\cite{tiradogarin2025anycalib}, on the three 3D reconstruction datasets.
The ScanNet++ calibration-only subset uses the undistorted mixed-device protocol rather than the mixed-projection protocol.
Table~\ref{tab:estimated_k}(a) measures the estimator's calibration accuracy independently of 3D reconstruction.
Relative focal error averages the relative errors of $f_x$ and $f_y$, while principal-point error is measured in pixels.

AnyCalib's average focal errors are 7.1\% on ETH3D and 8.6\% on ScanNet++, with larger focal and principal-point errors on PandaSet.
Unlike the uniform focal perturbations in Sec.~\ref{subsec:calibration_sensitivity}, estimated calibration contains coupled, image-dependent errors, motivating an end-to-end 3D reconstruction evaluation.

Table~\ref{tab:estimated_k}(b) reports 16-view \method{} 3D reconstruction when conditioning geometry is replaced by off-the-shelf estimates while ground-truth rays and depth remain the evaluation targets.
On undistorted mixed-device ScanNet++, we compare ground-truth intrinsics with estimates from single-image AnyCalib~\cite{tiradogarin2025anycalib} and multi-view VGGT~\cite{wang2025vggt}.
On the mixed-projection protocol, DSLR views retain their non-pinhole projection while iPhone views use undistorted pinhole images, and both the ground-truth and estimated conditioning geometry are represented as unit-ray maps.
AnyCalib uses the four-coefficient Kannala--Brandt model for non-pinhole DSLR views and the pinhole model for iPhone views.
Depth error uses $z$-depth for the undistorted protocol and ray depth for the mixed-projection protocol.

On undistorted mixed-device ScanNet++, VGGT-estimated intrinsics outperform single-image AnyCalib, with point relative error of $0.066$ versus $0.116$ and point inliers of $71.0\%$ versus $50.3\%$.
Both estimators remain less accurate than ground-truth calibration.
On the mixed-projection protocol, AnyCalib-estimated unit-ray maps yield a pointmap relative error of $0.088$, a ray-direction error of $2.16^\circ$, and point inliers of $60.2\%$, compared with $0.059$, $0.71^\circ$, and $72.9\%$ under ground-truth ray maps.
These results demonstrate the use of estimated image-to-ray geometry for non-pinhole 3D reconstruction while quantifying the remaining gap to ground-truth calibration.

\subsection{Resolution and Projection-Model Scope}
\label{app:resolution_scope}

The resolution evaluation in Fig.~\ref{fig:res_extrap} covers four-view ETH3D inputs up to $2\times$ the 518-pixel training long side, using calibrated pinhole images and digital resampling.
Mixed-projection ScanNet++ further evaluates non-pinhole DSLR and pinhole iPhone inputs with ray-depth metrics.
These results establish evidence for the tested resolutions and calibrated projection models, rather than general support for arbitrary non-pinhole systems.
Panoramic, catadioptric, and rolling-shutter imaging require appropriate image-to-ray geometry and separate end-to-end evaluation.

\subsection{Controlled FoV Variation Scope}
\label{app:fov_scope}

Objaverse varies FoV during rendering, while the real-image evaluations synthesize controlled FoV variation through cropping and resizing.
The aspect-preserving $256{\times}144$ RE10K evaluation avoids upsampling typical $640{\times}360$ source images within the tested zoom range.
These protocols do not capture changes in point-spread function, distortion, or exposure that may accompany optical zoom, so robustness to those effects remains to be evaluated.

\end{document}